%% file: main.tex
\documentclass{article}

\usepackage[preprint]{icml2026}

\usepackage{microtype}
\usepackage{graphicx}
\usepackage{subcaption}
\usepackage{booktabs}
\usepackage{float}
\usepackage{hyperref}
\usepackage{xurl}
\usepackage{tikz}
\usetikzlibrary{arrows.meta,positioning,calc,backgrounds}

\input{includes/base}
\editenable{} 
\input{includes/base_editing}
\input{includes/commands}

\input{includes/symbols}
\usepackage{arydshln}

\usepackage{amsmath}
\usepackage{amssymb}
\usepackage{mathtools}
\usepackage{amsthm}
\usepackage[capitalize,noabbrev]{cleveref}

\usepackage{graphicx}
\usepackage{multirow}
\usepackage{arydshln}
\usepackage[table]{xcolor}

\icmltitlerunning{MiSTiQ: MicroScaling Trellis Quantization for Extreme KV Cache Compression}

\begin{document}

\twocolumn[
  \icmltitle{MiSTiQ: MicroScaling Trellis Quantization for Extreme KV Cache Compression}
  \icmlsetsymbol{equallast}{*}
  \begin{icmlauthorlist}
  \icmlauthor{Ofir Gordon}{A}
  \icmlauthor{Lior Dikstein}{A}
  \icmlauthor{Paul Balanca}{G}
  \icmlauthor{Idan Achituve}{A}
  \icmlauthor{Douglas Orr}{equallast,G}
  \icmlauthor{Hai Victor Habi}{equallast,A}
  \end{icmlauthorlist}
  \icmlaffiliation{A}{Arm AI Research (AAIR), Arm Ltd}
  \icmlaffiliation{G}{Graphcore Research}
  \icmlcorrespondingauthor{Hai Victor Habi}{haivictor.habi@arm.com}
  \icmlcorrespondingauthor{Douglas Orr}{douglaso@graphcore.ai}
  \vskip 0.3in
]
\printAffiliationsAndNotice{\textsuperscript{*}Equal last authors. }
\input{section_files/abstract}
\input{files/introduction}
\input{section_files/related_work}
\input{section_files/figure1_overview}

\input{files/background}
\input{files/beam_search}

\input{files/methods}
\input{files/theortical_analysis}

\input{files/experiments}
\input{files/conclusion}
\clearpage
\bibliographystyle{icml2026}
\bibliography{ref_checked}

\clearpage
\appendix
\raggedbottom
\onecolumn
\input{section_files/implementation_details}
\input{section_files/alg_pseudo_code}

\input{files/proof}
\input{section_files/mx_gaussian_proof}

\input{section_files/clipped_mx_hash}
\input{section_files/efficient_lut}
\input{section_files/joint_scale_search}
\input{section_files/additional_results}

\end{document}

%% file: includes/base.tex
\usepackage{ifthen}
\newboolean{edit}
\setboolean{edit}{true}
\newcommand{\editenable}[1]{\setboolean{edit}{true}}
\newcommand{\editdisable}[1]{\setboolean{edit}{false}}

\usepackage{amsthm}
\usepackage{amssymb}
\usepackage{algorithm}
\usepackage{amsmath}
\usepackage{acronym}

\usepackage{commath,amsmath,amssymb,amsfonts}

\usepackage{accents}
\usepackage[inkscapearea=page]{svg}

\usepackage{mathtools}
\usepackage{hyperref}
\usepackage{amssymb}
\usepackage{bm}
\usepackage{graphicx}
\usepackage{float}
\usepackage{amsmath}
\usepackage{tikz}
\usepackage{multicol}
\usepackage{dsfont}
\usepackage{tcolorbox}
\usepackage{bbm}
\usepackage{multirow}
\usetikzlibrary{positioning, backgrounds, fit, shapes.arrows}
\usetikzlibrary{decorations.markings}
\usepackage{cuted}
\usepackage{enumitem}
\usepackage{caption}
\usepackage{booktabs}
\usepackage{longtable} 
\usepackage{subcaption}
\newtheorem{definition}{Definition}[section]
\newtheorem{theorem}{Theorem}[section]

\newtheorem{lemma}[theorem]{Lemma}

\usepackage{algpseudocode}
\tikzset{block/.style = {draw, fill=white, rectangle,
		minimum height=3em, minimum width=2cm},
	input/.style = {coordinate},
	output/.style = {coordinate},
	pinstyle/.style = {pin edge={to-,t,black}}
	radiation/.style={{decorate,decoration={expanding waves,angle=90,segment   length=4pt}}}
	
}
\usepackage{smartdiagram}

\tikzstyle{block} = [draw, rectangle, minimum height=2em, minimum width=2em]
\tikzstyle{sum} = [draw, circle,minimum width=0.1 cm]
\tikzstyle{input} = [coordinate]
\tikzstyle{output} = [coordinate]
\tikzstyle{dummy} = [coordinate]
\tikzstyle{pinstyle} = [pin edge={to-,thin,black}]
\usetikzlibrary{positioning, fit, arrows.meta}
\usetikzlibrary{positioning}
\usetikzlibrary{shapes,arrows}
\tikzstyle{frame_cyan} = [thick, draw=blue, solid,inner sep=0.3em]
\tikzstyle{frame_red} = [thick, draw=red, solid,inner sep=0.3em]
\tikzstyle{frame_green} = [thick, draw=green, solid,inner sep=0.3em]

\usepackage{tikz}
\usepackage{xcolor}
\definecolor{fc}{HTML}{1E90FF}
\definecolor{h}{HTML}{228B22}
\definecolor{bias}{HTML}{87CEFA}
\definecolor{noise}{HTML}{8B008B}
\definecolor{conv}{HTML}{FFA500}
\definecolor{pool}{HTML}{B22222}
\definecolor{up}{HTML}{B22222}
\definecolor{view}{HTML}{FFFFFF}
\definecolor{bn}{HTML}{FFD700}
\tikzset{fc/.style={black,draw=black,fill=fc,rectangle,minimum height=1cm}}
\tikzset{h/.style={black,draw=black,fill=h,rectangle,minimum height=1cm}}
\tikzset{bias/.style={black,draw=black,fill=bias,rectangle,minimum height=1cm}}
\tikzset{noise/.style={black,draw=black,fill=noise,rectangle,minimum height=1cm}}
\tikzset{conv/.style={black,draw=black,fill=conv,rectangle,minimum height=1cm}}
\tikzset{pool/.style={black,draw=black,fill=pool,rectangle,minimum height=1cm}}
\tikzset{up/.style={black,draw=black,fill=up,rectangle,minimum height=1cm}}
\tikzset{view/.style={black,draw=black,fill=view,rectangle,minimum height=1cm}}
\tikzset{bn/.style={black,draw=black,fill=bn,rectangle,minimum height=1cm}}
 
\usepackage{xspace}

\tikzstyle{dummy} = [coordinate]
\pgfkeys{/pgf/.cd,
  parallelepiped offset x/.initial=2mm,
  parallelepiped offset y/.initial=2mm
}
\pgfdeclareshape{parallelepiped}
{
  \inheritsavedanchors[from=rectangle] 
  \inheritanchorborder[from=rectangle]
  \inheritanchor[from=rectangle]{north}
  \inheritanchor[from=rectangle]{north west}
  \inheritanchor[from=rectangle]{north east}
  \inheritanchor[from=rectangle]{center}
  \inheritanchor[from=rectangle]{west}
  \inheritanchor[from=rectangle]{east}
  \inheritanchor[from=rectangle]{mid}
  \inheritanchor[from=rectangle]{mid west}
  \inheritanchor[from=rectangle]{mid east}
  \inheritanchor[from=rectangle]{base}
  \inheritanchor[from=rectangle]{base west}
  \inheritanchor[from=rectangle]{base east}
  \inheritanchor[from=rectangle]{south}
  \inheritanchor[from=rectangle]{south west}
  \inheritanchor[from=rectangle]{south east}
  \backgroundpath{
    \southwest \pgf@xa=\pgf@x \pgf@ya=\pgf@y
    \northeast \pgf@xb=\pgf@x \pgf@yb=\pgf@y
    \pgfmathsetlength\pgfutil@tempdima{\pgfkeysvalueof{/pgf/parallelepiped offset x}}
    \pgfmathsetlength\pgfutil@tempdimb{\pgfkeysvalueof{/pgf/parallelepiped offset y}}
    \def\ppd@offset{\pgfpoint{\pgfutil@tempdima}{\pgfutil@tempdimb}}
    \pgfpathmoveto{\pgfqpoint{\pgf@xa}{\pgf@ya}}
    \pgfpathlineto{\pgfqpoint{\pgf@xb}{\pgf@ya}}
    \pgfpathlineto{\pgfqpoint{\pgf@xb}{\pgf@yb}}
    \pgfpathlineto{\pgfqpoint{\pgf@xa}{\pgf@yb}}
    \pgfpathclose
    \pgfpathmoveto{\pgfqpoint{\pgf@xb}{\pgf@ya}}
    \pgfpathlineto{\pgfpointadd{\pgfpoint{\pgf@xb}{\pgf@ya}}{\ppd@offset}}
    \pgfpathlineto{\pgfpointadd{\pgfpoint{\pgf@xb}{\pgf@yb}}{\ppd@offset}}
    \pgfpathlineto{\pgfpointadd{\pgfpoint{\pgf@xa}{\pgf@yb}}{\ppd@offset}}
    \pgfpathlineto{\pgfqpoint{\pgf@xa}{\pgf@yb}}
    \pgfpathmoveto{\pgfqpoint{\pgf@xb}{\pgf@yb}}
    \pgfpathlineto{\pgfpointadd{\pgfpoint{\pgf@xb}{\pgf@yb}}{\ppd@offset}}
  }
}
\pgfdeclareshape{document}{
\inheritsavedanchors[from=rectangle] 
\inheritanchorborder[from=rectangle]
\inheritanchor[from=rectangle]{center}
\inheritanchor[from=rectangle]{north}
\inheritanchor[from=rectangle]{north east}
\inheritanchor[from=rectangle]{north west}
\inheritanchor[from=rectangle]{south}
\inheritanchor[from=rectangle]{south east}
\inheritanchor[from=rectangle]{south west}
\inheritanchor[from=rectangle]{west}
\inheritanchor[from=rectangle]{east}
\backgroundpath{%
\southwest \pgf@xa=\pgf@x \pgf@ya=\pgf@y
\northeast \pgf@xb=\pgf@x \pgf@yb=\pgf@y
\pgf@xc=\pgf@xb \advance\pgf@xc by-5pt 
\pgf@yc=\pgf@ya \advance\pgf@yc by5pt
\pgfpathmoveto{\pgfpoint{\pgf@xa}{\pgf@ya}}
\pgfpathlineto{\pgfpoint{\pgf@xa}{\pgf@yb}}
\pgfpathlineto{\pgfpoint{\pgf@xb}{\pgf@yb}}
\pgfpathlineto{\pgfpoint{\pgf@xb}{\pgf@yc}}
\pgfpathlineto{\pgfpoint{\pgf@xc}{\pgf@ya}}
\pgfpathclose
\pgfpathmoveto{\pgfpoint{\pgf@xc}{\pgf@ya}}
\pgfpathlineto{\pgfpoint{\pgf@xc}{\pgf@yc}}
\pgfpathlineto{\pgfpoint{\pgf@xb}{\pgf@yc}}
\pgfpathclose
}
}
\tikzstyle{block} = [draw, fill=white, rectangle, 
    minimum height=3em, minimum width=6em]
    
\usetikzlibrary{backgrounds}
\usepackage{pifont}
\tikzstyle{startstop} = [rectangle, rounded corners, minimum width=2cm, minimum height=0.7cm,text centered, draw=black, fill=lime!30]
\usepackage{stackengine}

\makeatletter
\newcounter{phase}[algorithm]
\newlength{\phaserulewidth}
\newcommand{\setphaserulewidth}{\setlength{\phaserulewidth}}

\makeatother
\setphaserulewidth{.7pt}
\makeatother
\usepackage{colortbl}

%% file: includes/base_editing.tex
\ifthenelse{\boolean{edit}}{%
  \usepackage{todonotes}
  \usepackage{color-edits} 
}{%
  \usepackage[disable]{todonotes}
  \usepackage[suppress]{color-edits} 
}

%% file: includes/commands.tex
\newcommand\MakeUppercaseGreek[1]{
  \begingroup
    \let\psi\Psi
    \let\omega\Omega
    \let\gamma\Gamma
    \MakeUppercase{#1}
  \endgroup}
\newcommand{\vectorsym}[1]{\bm{#1}}
\newcommand{\randomvec}[1]{\MakeUppercaseGreek{\bm{#1}}}

\newcommand{\matsym}[1]{\mathbf{#1}}

\newcommand{\probP}{\text{I\kern-0.15em P}}

%% file: includes/symbols.tex
\newcommand{\R}[0]{\mathbb{R}} 

\newcommand{\y}[0]{\vectorsym{y}}

\newcommand{\x}[0]{\vectorsym{x}}

\newcommand{\mxsignbit}{\gamma} 
\newcommand{\dmx}{d_{\mathrm{mx}}} 
\newcommand{\mxnumblocks}{N_B} 
\newcommand{\mxindexset}{\mathcal I} 
\newcommand{\mxrange}{C_S} 
\newcommand{\mxrefscale}{s^{*}} 
\newcommand{\mxscale}{g_{\mathrm{mxs}}} 
\newcommand{\stderf}{\mathrm{erf}} 
\newcommand{\stdierf}{\mathrm{ierf}} 
\newcommand{\hashparams}{\Omega} 
\newcommand{\lcg}{\operatorname{LCG}} 
\newcommand{\mxmix}{\operatorname{Mixer}} 
\newcommand{\mxhashvec}{\vectorsym{h}_{\hashparams}} 

\newcommand{\qtipmemory}{L} 
\newcommand{\qtiprate}{k} 
\newcommand{\qtipwidth}{V} 

\newcommand{\trellissource}{\mathbf r} 
\newcommand{\trellisrecon}{\widehat{\mathbf r}} 
\newcommand{\mxcellmass}{P} 
\newcommand{\heuristicscale}{s^{\mathrm{ref}}} 

\newcommand{\mxmixstep}{\operatorname{M}} 
\newcommand{\mxxorshift}{\operatorname{S}} 

\newcommand{\mxelementgrid}{\mathcal Q_E} 
\newcommand{\mxscalegrid}{\mathcal Q_S} 

\newcommand{\lutslopes}{c^{\mathrm{LUT}}} 
\newcommand{\lutoffset}{d^{\mathrm{LUT}}} 
\newcommand{\lutargument}{\eta} 
\newcommand{\lutknots}{z^{\mathrm{LUT}}} 
\newcommand{\lutvalues}{v^{\mathrm{LUT}}} 

\newcommand{\graphG}{\mathcal G} 

\newcommand{\beamstartbits}{b_{\mathrm{start}}} 
\newcommand{\beamset}{\mathcal B} 
\newcommand{\beamhyp}{h} 
\newcommand{\beamterminal}{u} 
\newcommand{\beamcost}{J} 
\newcommand{\beambackpointer}{\pi} 
\newcommand{\beamcandidates}{\mathcal C} 

\newcommand{\MN}{MiSTiQ}
\newcommand{\parameffrate}{b_{\mathrm{eff}}} 

\newcommand{\TableMN}{\textbf{\MN{} (Ours)}}

\newcommand{\clipradius}{R} 
\newcommand{\clipoperator}{\operatorname{clip}} 
\newcommand{\clippedcoord}{\widetilde X} 
\newcommand{\clippedcdf}{F_{\mathrm{clip}}} 
\newcommand{\clipcross}{p_{\mathrm{clip}}^\star} 
\newcommand{\clippedhash}{h^{\mathrm{clip}}} 


%% file: section_files/abstract.tex
\begin{abstract}
Modern large language models (LLMs) increasingly support longer contexts and serve more concurrent requests, increasing both sequence lengths and batch sizes.
Both trends increase KV-cache size and memory traffic during inference.
KV-cache quantization reduces these costs, with recent approaches using codebook-based vector quantization to achieve extremely low bit widths.
However, hardware constraints limit codebook size, and most methods reconstruct the cache in higher-precision floating-point formats for attention, limiting the computational benefits of quantization.
To address these limitations, we introduce MiSTiQ, a drop-in replacement for codebook-based vector quantizers that combines trellis-coded storage with microscaling (MX) formats for low-precision computation.
We combine trellis quantization with beam search to encode the KV cache over large reconstruction spaces without large codebooks. 
To support MX computation, we derive MX-Hash functions that generate reconstruction values directly in the target MX element format. These functions use the scale-conditioned marginal distribution of block-normalized Gaussian inputs.
On Llama3 and Qwen3, MiSTiQ achieves lower WikiText perplexity and higher GSM8K accuracy than the evaluated baselines across several block-scale formats at approximately 1.25 effective bits per element. With NVFP4, it also achieves the highest LongBench average on both models.
\end{abstract}

%% file: files/introduction.tex
\section{Introduction}

Large language models (LLMs) have been broadly adopted across many applications \cite{brown2020language,taylor2022galactica,yuan2024large}. In many cases, these applications require long-context inputs, such as long-context reasoning \cite{wei2022chain,guo2025deepseek} and retrieval-augmented generation (RAG) \cite{lewis2020retrieval}. However, handling long contexts during LLM inference demands substantial memory capacity and incurs high memory traffic due to key-value cache. Moreover, in certain scenarios—such as when using speculative decoding \cite{flashinfer}, with Multi-head Latent Attention (MLA) \cite{yuan2026blasst}, or when multiple requests share a common prefix \cite{juravsky2024hydragen} the decode phase can also become highly compute-intensive and may even be compute-bound.

To tackle these challenges, there are two main ways to approach the problem. The first is through architectural changes, such as Multi-query attention (MQA), Grouped-query attention (GQA) \cite{ainslie-etal-2023-gqa}, Sliding-window attention (SWA) \cite{team2025gemma}, and hybrid local \& global attention\cite{team2025gemma}. The second approach is to apply post-training compression techniques, including quantization \cite{liu2024kivi,hooper2024kvquant}, token selection \cite{zhang2023h2o, li2024snapkv}, and others. 
In this work, we focus on quantization for its potential to reduce memory, as well as its applicability to pretrained models without requiring extensive retraining.

KV-cache quantization methods exploit the different distributions of keys and values to reduce storage while preserving accuracy.
KIVI~\cite{liu2024kivi} uses different quantization granularities for keys and values, while more recent methods adopt vector quantization to improve accuracy at lower storage budgets.
These include NSNQuant~\cite{son2025nsnquant}, which uses calibration-free distribution shaping through normalization and scaling, and VecInfer~\cite{yao2026vecinfer}, which applies smoothing and Hadamard transformations to mitigate key-cache outliers.
These methods store the quantized cache as integer codes and use floating-point or integer arithmetic for computation.
However, codebook-based vector quantization is constrained by the size of the codebooks that hardware can efficiently support.
QTIP~\cite{tseng2025qtip} addresses this limitation for weight quantization through trellis quantization, enabling large reconstruction spaces without explicitly storing large codebooks.

Recently, the Open Compute Platform introduced the Microscaling data format \cite{rouhani2023mx}, a collaboration among Arm, AMD, NVIDIA, Qualcomm, Intel, Meta, and Microsoft, aimed at standardizing computation for deep learning models. 
Microscaling (MX) formats have attracted substantial attention in research on weight and activation quantization, including work on GPTQ adaptation \cite{egiazarian2025mrgptq}, feature map rotation \cite{shao2025block,gordon2026latmix}, block-scale format representation \cite{fasoli2026finer}, and other efforts. 
While prior studies have explored MX-based KV-cache quantization at four bits or above~\citep{lee2025mx,zhang2026benchmark}, combining lower storage costs with direct reconstruction into MX-compatible formats remains an open challenge.

Here, we aim to reduce KV-cache storage while enabling direct reconstruction into MX-compatible formats for low-precision computation.
To achieve this, we address the codebook-size constraint by adapting trellis quantization for KV-cache compression. 
We then augment trellis quantization with a microscaling format to gain further computational benefits. 
The resulting approach is a drop-in quantizer that can replace vector quantization and be integrated with most methods. 
Concretely, we introduce Microscaling Trellis Quantization (\MN{}), which replaces QTIP's Viterbi encoder with beam search and uses MX-Hash functions based on the distribution of block-normalized Gaussian elements.
Furthermore, we incorporate joint scale optimization into the beam search with the finer-resolution UE5M3~\cite{fasoli2026finer} scale format.
To the best of our knowledge, MiSTiQ is the first dedicated KV-cache quantizer to combine sub-2-bit effective storage with direct reconstruction into MX-compatible formats.
Experiments on WikiText, LongBench, and GSM8K demonstrate improved accuracy over the evaluated baselines across a range of models and quantization settings.

\noindent\textbf{Contributions:}\enspace
We investigate approaches and tradeoffs for using QTIP as a storage format while leveraging MX formats for computation. 
This paper makes the following contributions: (1) we introduce beam search for trellis quantization, enabling trellis quantization of the key–value cache; 
(2) we develop QTIP hash variants tailored to match the marginal distribution of block-normalized elements; 
(3) we analyze design-parameter tradeoffs and evaluate the approach for KV-cache compression; and 
(4) we report experimental results across multiple benchmarks and models that demonstrate the benefits of the proposed method.
To support reproducibility, we will publicly release the code implementing our method upon acceptance. The appendix is provided separately as supplementary material.

%% file: section_files/related_work.tex
\section{Related Work}
\label{sec:related_work}

\paragraph{KV-cache compression.}
The growing memory and bandwidth costs of KV caches under long
contexts and large batches have driven extensive research into
cache compression.
Among these approaches, quantization has been widely studied
as a means of reducing the storage cost per cached element
while retaining the cached tokens.
KIVI uses per-channel key and per-token value
quantization~\citep{liu2024kivi}, while KVQuant combines pre-RoPE
key quantization with sensitivity-weighted nonuniform levels and outlier handling~\citep{hooper2024kvquant}.
Rotation-based approaches, including QuaRot, RotateKV, and TurboQuant, reshape the quantization domain to improve low-bit
representations~\citep{ashkboos2024quarot,rotatekv2025,zandieh2025turboquant}.
Codebook-based vector quantization instead exploits dependencies
across coordinates, using coupled channels or additive representations~\citep{zhang2024cq,li2025commvq}.
NSNQuant specifically combines double normalization, channel centering, and
Hadamard transforms to support a reusable Gaussian codebook without data-dependent calibration~\citep{son2025nsnquant}.
VecInfer combines smoothing and Hadamard transforms to improve
codebook utilization, with fused dequantization and attention
computation~\citep{yao2026vecinfer}.
Motivated by differences in quantization sensitivity, particularly
the greater sensitivity of keys, mixed-precision methods allocate
bits according to K/V and layer
sensitivity~\citep{pmlr-v267-li25dd,tao-etal-2025-asymkv},
token saliency~\citep{dong2024qaq,he2024zipcache}, or recency and progressive quantization
schedules~\citep{duanmu2024skvq,liu2026pm}.
Complementary approaches like AQUA-KV exploit reconstruction structure: quantizes residuals from learned predictors~\citep{shutova2025cache}.
Other compression mechanisms include token selection and eviction~\citep{zhang2023h2o,li2024snapkv}, head-dependent cache retention~\citep{xiao2025duoattention}, and low-rank cache representations~\citep{chang2025palu}.

\paragraph{Trellis quantization.}
Trellis-coded quantization represents source sequences through paths in a finite-state trellis, with subsequent extensions assigning vector reconstructions to trellis branches~\citep{marcellin1990tcq,fischer1991tcvq}.
This structure couples successive reconstruction choices while permitting encoding by dynamic programming, avoiding an explicit enumeration of all possible reconstructed sequences.
For low-bit LLM weight compression, QuIP\# combines randomized Hadamard preprocessing with $E_8$ lattice codebooks at up to four bits per weight~\citep{tseng2024quip}, while AQLM uses sums of learned codewords with joint optimization within transformer blocks, targeting two to three bits per parameter~\citep{egiazarian2024aqlm}.
However, the size of an explicit, unstructured codebook grows exponentially with vector dimension at a fixed bitrate.
QTIP combines trellis quantization with incoherence processing, using a stateful bitshift trellis to decouple the effective quantization dimension from the stored codebook size~\citep{tseng2025qtip}.
It supports both lookup-based and computed, lookup-free reconstruction rules, offering different tradeoffs between storage and computation.
Proteus further develops lookup-free trellis quantization through computed reconstruction codes~\citep{yang2026proteus}.


\paragraph{MX quantization.}

Microscaling (MX) formats combine low-precision floating-point or integer elements with a shared block scale, extending their representable dynamic range~\citep{rouhani2023mx}.
Early work adapted established post-training quantization algorithms to MX formats~\citep{sharify2024post}, while subsequent studies examined the effects of element format and quantization granularity~\citep{chen2026intvsfp}.
Format extensions such as MX+ and AMXFP4 improve outlier representation through enhanced element precision and asymmetric shared scales, respectively~\citep{lee2025mx,lee2025amxfp4}.
Rotation-based approaches, including MR-GPTQ and Block Rotation, use block-local transformations tailored to MX scaling constraints~\citep{egiazarian2025mrgptq,shao2025block}.
FPTQuant also reshapes activations and improves weight--activation alignment for low-bit integer quantization~\citep{breugel2026fptquant}.
For MX quantization, LATMiX extends distribution shaping beyond rotations through learned invertible affine transformations that account for the activation distribution and MX block structure~\citep{gordon2026latmix}.
Complementary approaches improve scaling through pre-scale optimization~\citep{zhang2026benchmark}, overflow-aware scale selection, and higher-precision scales shared across larger blocks~\citep{chhugani2026scaling}.


%% file: section_files/figure1_overview.tex
\begin{figure*}[!t]
  \centering
  \resizebox{0.94\textwidth}{!}{\input{figures/figure1_overview.tikz}}
  \caption{\textbf{MicroScaling trellis quantization overview.}
  (a) Inputs are scaled blockwise before beam search on the bitshift trellis.
  (b) Trellis-state bits are mixed and mapped through the scale-dependent inverse CDF to form MX-Gaussian reconstruction values. The distribution diagrams are schematic.}
  \label{fig:mistiq-overview}
\end{figure*}
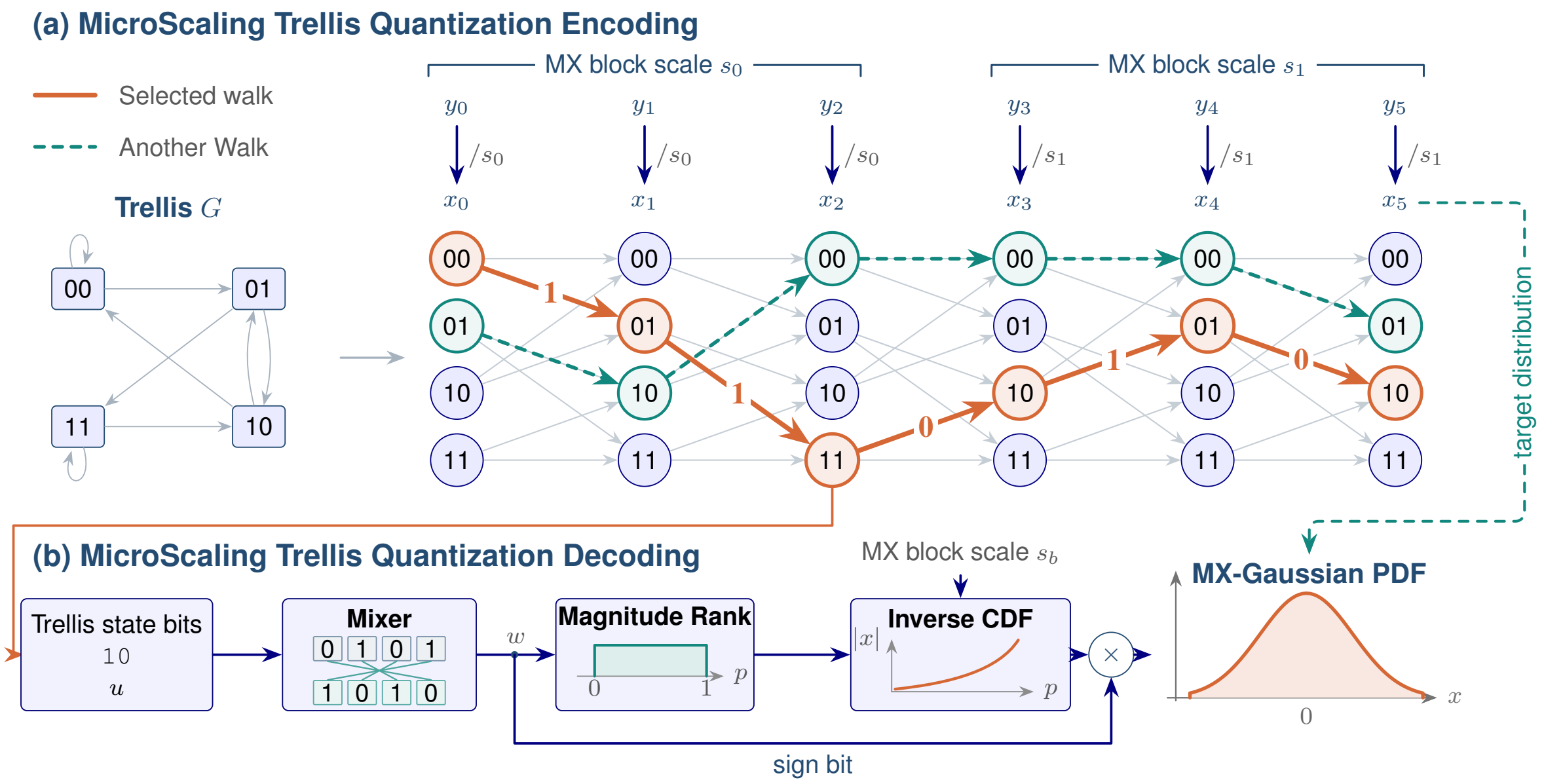

%% file: figures/figure1_overview.tikz
\begin{tikzpicture}[x=1cm,y=1cm,font=\sffamily\fontsize{8.3}{10.3}\selectfont,
  >=Stealth, line cap=round, line join=round,
  state/.style={circle,draw=blue!50!black,fill=blue!8,minimum size=6.1mm,inner sep=0pt},
  block/.style={rounded corners=2pt,draw=blue!45!black,fill=blue!5,minimum height=13mm,align=center},
  flow/.style={->,line width=.8pt,draw=blue!50!black}]
\definecolor{mxblue}{HTML}{244C72}
\definecolor{mxorange}{HTML}{D66836}
\definecolor{mxteal}{HTML}{12877E}
\definecolor{mxgray}{HTML}{A6B2BE}
\path[use as bounding box] (-.55,-.3) rectangle (17.9,8.65);
\node[anchor=west,font=\sffamily\bfseries\fontsize{10}{12}\selectfont,text=mxblue] at (0,8.35) {(a) MicroScaling Trellis Quantization Encoding};
\draw[mxorange,line width=1.5pt] (.15,7.55) -- (.85,7.55);
\node[anchor=west,text=black!65] at (1.0,7.55) {Selected walk};
\draw[mxteal,dashed,line width=1.2pt] (.15,6.95) -- (.85,6.95);
\node[anchor=west,text=black!65] at (1.0,6.95) {Another Walk};
\draw[mxblue,line width=.65pt] (4.72,7.77) -- (4.72,7.9) -- (9.74,7.9) -- (9.74,7.77);
\node[fill=white,inner sep=3pt,text=mxblue] at (7.23,7.9) {MX block scale $s_0$};
\draw[mxblue,line width=.65pt] (11.26,7.77) -- (11.26,7.9) -- (16.28,7.9) -- (16.28,7.77);
\node[fill=white,inner sep=3pt,text=mxblue] at (13.77,7.9) {MX block scale $s_1$};
\node[font=\sffamily\bfseries\fontsize{9}{11}\selectfont,text=mxblue] at (1.7,6.25) {Trellis $G$};
\begin{scope}[state/.style={rectangle,rounded corners=1.5pt,draw=mxblue,fill=blue!8,minimum width=6.1mm,minimum height=4.8mm,inner sep=0pt}]
\node[state] (g0) at (.65,5.3) {00};
\node[state] (g1) at (2.75,5.3) {01};
\node[state] (g2) at (2.75,3.7) {10};
\node[state] (g3) at (.65,3.7) {11};
\end{scope}
\draw[->,mxgray] (g0) edge[loop above] (g0);
\draw[->,mxgray] (g3) edge[loop below] (g3);
\draw[->,mxgray] (g0) -- (g1);
\draw[->,mxgray] (g1) -- (g3);
\draw[->,mxgray] (g3) -- (g2);
\draw[->,mxgray] (g2) -- (g0);
\draw[->,mxgray] (g1) edge[bend left=13] (g2);
\draw[->,mxgray] (g2) edge[bend left=13] (g1);
\draw[->,mxgray,line width=.7pt] (3.7,4.5) -- (4.45,4.5);
\foreach \t in {0,...,5}{
  \foreach \u/\bits in {0/00,1/01,2/10,3/11}{
    \node[state] (s\t\u) at ({5.05+2.18*\t},{5.65-.78*\u}) {\bits};
  }
  \node[text=mxblue] (y\t) at ({5.05+2.18*\t},7.4) {$y_{\t}$};
  \node[text=mxblue] (x\t) at ({5.05+2.18*\t},6.3) {$x_{\t}$};
  \pgfmathtruncatemacro{\b}{int(\t/3)}
  \draw[flow] (y\t.south) -- node[right,text=black!65] {$/s_{\b}$} (x\t.north);
}
\begin{scope}[on background layer]
\foreach \t [evaluate=\t as \next using int(\t+1)] in {0,...,4}{
 \foreach \u in {0,...,3}{
  \foreach \c in {0,1}{
   \pgfmathtruncatemacro{\v}{mod(2*\u+\c,4)}
   \draw[->,mxgray!65,line width=.45pt] (s\t\u) -- (s\next\v);
  }
 }
}
\end{scope}
\foreach \t/\u/\next/\v in {0/1/1/2,1/2/2/0,2/0/3/0,3/0/4/0,4/0/5/1}{
 \draw[->,mxteal,dashed,line width=1.2pt] (s\t\u) -- (s\next\v);
}
\foreach \t/\u in {0/1,1/2,2/0,3/0,4/0,5/1}{
 \node[state,draw=mxteal,line width=.9pt,fill=mxteal!7] at (s\t\u) {\ifcase\u\relax 00\or01\or10\or11\fi};
}
\foreach \t/\u/\next/\v/\bit in {0/0/1/1/1,1/1/2/3/1,2/3/3/2/0,3/2/4/1/1,4/1/5/2/0}{
 \draw[->,mxorange,line width=1.6pt] (s\t\u) -- node[pos=.5,fill=white,inner sep=1.5pt,text=mxorange,font=\bfseries] {\bit} (s\next\v);
}
\foreach \t/\u in {0/0,1/1,2/3,3/2,4/1,5/2}{
 \node[state,draw=mxorange,line width=1.1pt,fill=mxorange!12] at (s\t\u) {\ifcase\u\relax 00\or01\or10\or11\fi};
}
\draw[flow,mxorange] (s23.south) -- (9.41,2.55) -- (-.1,2.55) -- (-.1,1.05) -- (.025,1.05);
\node[anchor=west,font=\sffamily\bfseries\fontsize{10}{12}\selectfont,text=mxblue] at (0,2.17) {(b) MicroScaling Trellis Quantization Decoding};
\node[block,minimum width=2.15cm] (bits) at (1.1,1.05) {Trellis state bits\\\texttt{10}\\$u$};
\node[block,minimum width=2.25cm] (mix) at (4.15,1.05) {};
\node[font=\sffamily\bfseries\fontsize{8.3}{10.3}\selectfont] at (4.15,1.47) {Mixer};
\foreach \j/\inbit/\outbit in {0/0/1,1/1/0,2/0/1,3/1/0}{
 \node[draw=mxblue!60,fill=mxblue!8,rounded corners=.4pt,minimum width=3.3mm,minimum height=2.8mm,inner sep=0pt] (in\j) at ({3.55+.4*\j},1.12) {\inbit};
 \node[draw=mxteal!70,fill=mxteal!9,rounded corners=.4pt,minimum width=3.3mm,minimum height=2.8mm,inner sep=0pt] (out\j) at ({3.55+.4*\j},.61) {\outbit};
}
\foreach \j/\k in {0/3,1/2,2/1,3/0}{
 \draw[mxteal!75,line width=.5pt] (in\j.south) -- (out\k.north);
}
\draw[flow] (bits) -- (mix);
\node[block,minimum width=2.3cm] (uniform) at (7.35,1.05) {};
\node[font=\sffamily\bfseries\fontsize{8.3}{10.3}\selectfont] at (7.35,1.47) {Magnitude Rank};
\path[fill=mxteal!14] (6.65,.8) rectangle (7.95,1.16);
\draw[->,black!45,line width=.4pt] (6.45,.8) -- (8.15,.8) node[right,text=black!65] {$p$};
\draw[mxteal,line width=.9pt] (6.65,.8) -- (6.65,1.16) -- (7.95,1.16) -- (7.95,.8);
\node[below,inner sep=1pt,text=black!65] at (6.65,.8) {$0$};
\node[below,inner sep=1pt,text=black!65] at (7.95,.8) {$1$};
\draw[flow] (mix) -- node[above,text=black!65] {$w$} (uniform);
\fill[mxblue] (5.72,1.05) circle (1.2pt);
\draw[flow] (5.72,1.05) -- (5.72,.02) -- node[below,text=mxblue] {sign bit} (12.65,.02) -- (12.65,.79);
\node[block,minimum width=2.55cm] (inverse) at (10.9,1.05) {};
\node[font=\sffamily\bfseries\fontsize{8.3}{10.3}\selectfont] at (10.9,1.47) {Inverse CDF};
\draw[->,black!45,line width=.4pt] (10.1,.62) -- (11.75,.62) node[right,text=black!65] {$p$};
\draw[->,black!45,line width=.4pt] (10.1,.62) -- (10.1,1.23) node[left,text=black!65] {$|x|$};
\draw[mxorange,line width=.95pt] (10.14,.65) .. controls (10.8,.72) and (11.34,.87) .. (11.57,1.22);
\draw[flow] (uniform) -- (inverse);
\node[text=black!65,anchor=south] at (10.9,1.98) {MX block scale $s_b$};
\draw[flow] (10.9,1.97) -- (inverse.north);
\node[circle,draw=mxblue,fill=white,minimum size=5.2mm,inner sep=0pt,text=mxblue] (signmul) at (12.65,1.05) {$\times$};
\draw[flow] (inverse.east) -- (signmul.west);
\draw[flow] (signmul.east) -- (13.12,1.05);
\begin{scope}[shift={(13.4,.55)},x=.82cm,y=.82cm]
 \draw[->,black!55,line width=.5pt] (-.12,0) -- (3.7,0) node[right] {$x$};
 \draw[->,black!55,line width=.5pt] (0,-.04) -- (0,1.8);
 \path[fill=mxorange!15] (.18,0) -- plot[domain=-1.65:1.65,samples=65] ({1.85+\x},{1.48*exp(-1.15*\x*\x)}) -- (3.5,0) -- cycle;
 \draw[mxorange,line width=1.15pt] (.2,0) -- (.2,.064) plot[domain=-1.65:1.65,samples=65] ({1.85+\x},{1.48*exp(-1.15*\x*\x)}) -- (3.5,0);
 \node[below,text=black!60] at (1.85,0) {$0$};
\end{scope}
\node[align=center,font=\sffamily\bfseries\fontsize{9}{11}\selectfont,text=mxblue] at (14.95,2.0) {MX-Gaussian PDF};
\draw[->,mxteal,dashed,line width=.9pt,rounded corners=3pt] (x5.east) -- (17.45,6.3) -- (17.45,2.6) -- (14.95,2.6) -- (14.95,2.21);
\node[rotate=90,text=mxteal,fill=white,inner sep=2pt] at (17.45,4.45) {target distribution};
\end{tikzpicture}

%% file: files/background.tex
\section{Background}
\subsection{Microscaling Quantization}
\label{sec:mx-background}
Microscaling represents a vector as blocks of low-precision elements, each
multiplied by a shared scale \citep{rouhani2023mx}.
Let $\y\in\mathbb R^d$ and assume that $d$ is a multiple of $\dmx$.
Define $\mxnumblocks=d/\dmx$ and
$\mxindexset_b=\{b\,\dmx,\ldots,(b+1)\dmx-1\}$ for
$b\in\{0,\ldots,\mxnumblocks-1\}$. We write $\y_b\in\mathbb R^{\dmx}$
for the coordinates indexed by $\mxindexset_b$. 
Let $\mxscalegrid$ and $\mxelementgrid$ denote the
representable scale and element point sets. The scale quantizer is
$Q_\alpha:\mathbb R\to\mxscalegrid$, and the element quantizer is
$Q_E:\mathbb R\to\mxelementgrid$. For a positive element-range constant $\mxrange$, a
common max-based reference rule and the resulting reconstruction are
\begin{equation}
\begin{aligned}
\mxrefscale_b &= \frac{\max_{i\in\mxindexset_b}|y_i|}{\mxrange},
&s_b &= Q_\alpha(\mxrefscale_b)\in\mxscalegrid,\\
\widehat y_i
&=s_b Q_E\!\left(y_i/s_b\right),
&& i\in\mxindexset_b .
\end{aligned}
\end{equation}
Here $Q_E$ acts elementwise.\footnote{The reconstruction formula assumes
$s_b>0$; if $s_b=0$, we set $\widehat y_i=0$ for all
$i\in\mxindexset_b$.}


The scale format determines the available range and precision of
$s_b$, while the element format determines the values available
after normalization. Scale quantization therefore contributes error
separately from element quantization. In particular, a limited scale
range can make smaller blocks less accurate when their reference
scales become unrepresentable \citep{fasoli2026finer}.

In this study, we employ two block-scale formats: (1) UE8M0, a power-of-two scaling format introduced in the microscaling framework \cite{rouhani2023mx}; and (2) UE5M3, an unsigned floating-point scale format with three mantissa bits and five exponent bits, introduced in \cite{fasoli2026finer}.


\subsection{QTIP}
\label{sec:qtip-background}
Vector quantization (VQ) represents a group of $\qtipwidth$ values by
one index selecting a reconstruction vector from a codebook. Jointly
chosen reconstruction vectors can provide better shaping than
independent scalar quantization. At $\qtiprate$ bits per element,
however, an unrestricted $\qtipwidth$-dimensional codebook contains
$2^{\qtiprate\qtipwidth}$ entries, so its storage and nearest-neighbor
search cost grow exponentially with $\qtipwidth$.

Trellis quantization represents long vectors that would be intractable
with an unrestricted codebook as a walk through a trellis.
QTIP \citep{tseng2025qtip} uses a $(\qtipmemory,\qtiprate,\qtipwidth)$ trellis,
where $\qtipmemory$ is the state-bit length, $\qtiprate$ is the
coded bit rate per scalar, and $\qtipwidth$ is the number of scalars
emitted at a trellis step. A length-$T$ source is divided into $n=T/V$ groups
$\trellissource_t\in\mathbb R^V$ for $t=1,\ldots,n$, assuming $V\mid T$.
The state set is $\mathcal U=\{0,\ldots,2^L-1\}$.
The reconstruction map $C:\mathcal U\to\mathbb R^V$ assigns a code value
to each state. It may use a stored codebook lookup or a computed QTIP code.

For predecessor $u\in\mathcal U$ and successor $v\in\mathcal U$,
$kV$ is an integer satisfying $1\le kV\le L$.
The bitshift edge relation and successor set are
\begin{equation}
\begin{aligned}
    (u,v)\in\mathcal{A}
    &\quad\Longleftrightarrow\quad
    \exists\,c\in\mathbb{Z},\;0\leq c<2^{kV}: \\
    &\qquad v=\left(u2^{kV}\bmod 2^L\right)+c, \\
    \operatorname{succ}(u)
    &=\left\{\left(u2^{kV}\bmod 2^L\right)+c:\right. \\
    &\qquad\left.c\in\mathbb{Z},\;0\leq c<2^{kV}\right\}.
\end{aligned}
\label{eq:bitshift_trellis_transition}
\end{equation}
That is, the most significant $L-kV$ bits of $v$ equal the least significant $L-kV$ bits of $u$.
Each transition appends $kV$ bits, so a state's output can be recovered
from a fixed-width window of the packed path bits. This permits parallel
decoding of output groups.

A valid state sequence $u_{1:n}$ reconstructs each group as $C(u_t)$.
Under squared error, define the local distortion and trellis objective by
\begin{equation}
\ell_t(u)=\|C(u)-\trellissource_t\|_2^2,
\qquad
\min_{\text{valid }u_{1:n}}D(u_{1:n}).
\label{eq:trellis-local-distortion}
\end{equation}
where $D(u_{1:n})=\sum_{t=1}^{n}\ell_t(u_t)$ is the total squared
reconstruction distortion of the trellis path $u_{1:n}$.
The additive cost permits Viterbi dynamic programming over trellis
states rather than enumeration of all reconstruction sequences.

QTIP applies incoherence processing to model weights before trellis
quantization, making the transformed source approximately
Gaussian-like. Its computed codes are designed to produce
approximately Gaussian outputs suited to that source model.

%% file: files/beam_search.tex
\section{Quantization with Beam Search on Trellises}
\label{sec:beam_search}

To enable trellis quantization of the KV cache while limiting encoding
overhead, we replace the exact Viterbi search used by QTIP
with approximate beam search.

\input{section_files/parameter_analysis_figure}

Let $\beamstartbits$ denote the number of bits used to select the initial
trellis state. Beam search considers initial states in
$\{0,\ldots,2^{\beamstartbits}-1\}$ and retains at most $B\geq1$ path
hypotheses after each expansion.
Let $\beamset_t$ be the indexed collection of retained hypotheses after
step $t$. Each hypothesis $\beamhyp\in\beamset_t$ has its own terminal state
$\beamterminal_t(\beamhyp)\in\mathcal U$, accumulated distortion
$\beamcost_t(\beamhyp)$, and predecessor-hypothesis index.
Initialize $\beamset_0=\{0,\ldots,2^{\beamstartbits}-1\}$, with
$\beamterminal_0(\beamhyp)=\beamhyp$ and $\beamcost_0(\beamhyp)=0$ for each
$\beamhyp\in\beamset_0$. Setting $\beamstartbits=0$ gives the fixed initial
state $u_0=0$. For each retained hypothesis
$\beamhyp\in\beamset_{t-1}$, expand every legal successor
$v\in\operatorname{succ}(\beamterminal_{t-1}(\beamhyp))$ with cost
\begin{equation}
    \widetilde{\beamcost}_t(\beamhyp,v)
    =\beamcost_{t-1}(\beamhyp)+\ell_t(v).
    \label{eq:beam_expansion}
\end{equation}
Keep up to $B$ lowest-cost expanded hypotheses across all parents and
assign them indices in $\beamset_t$.
At termination, select a minimum-cost retained
hypothesis and trace its predecessor indices to recover its path.
\Cref{alg:trellis_beam_encoding} includes a summary of the encoding procedure.

The state sequence \(u_{1:n}\) is the logical encoder output.  Its compact
stored representation comprises the branch labels
\(c_t=u_t\bmod 2^{kV}\), each occupying \(kV\) bits, together with the
selected initial state \(u_0\), represented by \(\beamstartbits\) bits.
The initial-state selector contributes \(\beamstartbits/T\) bits per scalar.
When \(\beamstartbits=0\), the decoder uses \(u_0=0\) without storing a selector.
Decoding consists only of state recovery, reconstruction-map lookup, and concatenation, without beam search or optimization.
The procedure is summarized in \cref{alg:trellis_direct_decoding}.

QTIP's optimized Viterbi search considers up to $2^L$ trellis states and has
complexity $O(2^L T/V)$. Beam search expands $2^{\beamstartbits}$ initial
hypotheses and at most $B$ hypotheses at each subsequent step, scoring
their $2^{kV}$ successors and requiring
$O((2^{\beamstartbits}+(T/V-1)B)2^{kV})$ work for direct distortion
evaluation, plus selection. This reduces the retained
search space at the cost of losing guaranteed optimality, while decoding
remains unchanged.


%% file: section_files/parameter_analysis_figure.tex
\begin{figure*}[!t]
  \centering
  \begin{subfigure}[t]{2.2in}
    \centering
    \includegraphics[width=\linewidth]{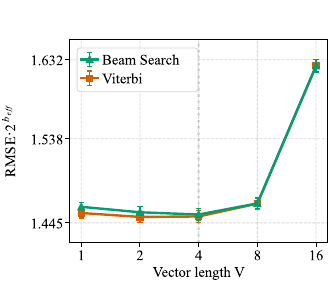}
    \caption{Vector length $\qtipwidth$}
    \label{fig:parameter-vector-length}
  \end{subfigure}\hspace{0.075in}%
  \begin{subfigure}[t]{2.2in}
    \centering
    \includegraphics[width=1\linewidth]{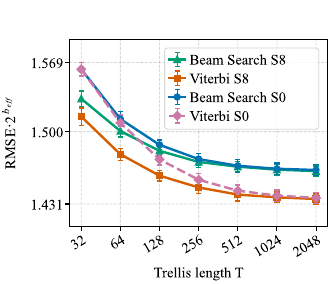}
    \caption{Trellis length $T$}
    \label{fig:parameter-trellis-length}
  \end{subfigure}\hspace{0.075in}%
  \begin{subfigure}[t]{2.2in}
    \centering
    \includegraphics[width=1\linewidth]{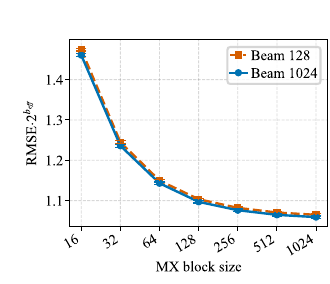}
    \caption{MX block size}
    \label{fig:parameter-mx-block-size}
  \end{subfigure}
  \caption{\textbf{MiSTiQ parameter analysis with standard Gaussian inputs}, consistent with the MX Gaussian hash derivation.
  (a) Vector length: beam search and Viterbi.
  (b) Trellis length: both methods with $\beamstartbits=8$ (S8) or $\beamstartbits=0$ (S0).
  (c) MX block size with fixed UE8M0 scales, $T=1024$, $\qtipwidth=8$, and beam widths 128 and 1024, at approximately matched effective rates.
  The ordinate is $\mathrm{RMSE}\cdot2^{\parameffrate}$; $\parameffrate$ counts payload, scale, and start-state bits per scalar.}
  \label{fig:parameter-analysis}
\end{figure*}

%% file: files/methods.tex
\section{MX Trellis Quantization (\MN{})}
\label{sec:mx-methods}

\begin{figure}
  \centering
  \includegraphics[width=1\linewidth]{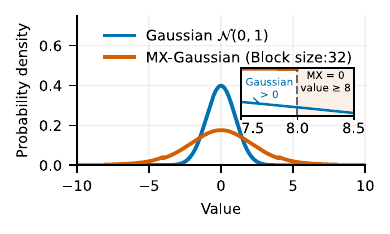}
  \caption{Comparison between the standard Gaussian distribution and the MX-Gaussian marginal distribution for a block size of 32 with E2M1 scaling. }
  \label{fig:gaussian-vs-mx-gaussian}
\end{figure}

We present a microscaling (MX) trellis quantization technique that integrates MX compute formats with a trellis-based storage scheme. We find that the ordering of scaling and trellis quantization affects reconstruction error. Instead of applying MX after trellis quantization, we propose performing MX scaling first and then encoding the scaled values into the trellis. Encoding these scaled values with the trellis requires a modified hash function that matches the Gaussian distribution after it has been scaled by the MX block scale; we refer to this scaled distribution as MX-Gaussian. In Figure ~\ref{fig:gaussian-vs-mx-gaussian} we illustrate the difference between Gaussian and MX-Gaussian. In addition, we incorporate MX element-wise quantization into the hash function so that the trellis beam search is explicitly aware of the element quantization. Finally, when using finer block-scale formats such as UE5M3 and NVFP4's E4M3 in place of UE8M0, we introduce a joint trellis and scale search to reduce reconstruction error.

We now formulate MX trellis quantization. Given a target vector $\y\in\mathbb R^d$, where $d=T$ is the scalar
length associated with the trellis, the goal is to obtain represented MX
block scales and a trellis code for its quantized reconstruction.
The trellis uses $n=T/V$ steps, each emitting $V$ coordinates\footnote{For this block-aligned construction, assume $V$ divides $\dmx$, so each
$V$-coordinate emission lies within one MX block.}.
MX trellis quantization combines trellis-coded storage with microscaling
computation \citep{rouhani2023mx}. Each MX block contains low-precision
elements and a shared scale, allowing dot products to be accumulated before
applying the product of the block scales.

Using the target blocks $\y_b$, block size $\dmx$, and quantizers defined
in \cref{sec:mx-background}, reconstruction takes the form
$\widehat{\y}_b=s_b\widehat{\x}_b$, where
$s_b\in\mxscalegrid$ and
$\widehat{\x}_b\in\mxelementgrid^{\dmx}$.
The normalized target is
$\x_b=\y_b/s_b$ for $s_b>0$.
The reconstruction error is measured in the original domain:
\begin{equation}
\|\y_b-\widehat{\y}_b\|_2^2
=\|\y_b-s_b\widehat{\x}_b\|_2^2
=s_b^2\|\x_b-\widehat{\x}_b\|_2^2,
\label{eq:mx-reconstruction-error}
\end{equation}
where $\y_b$ represent the $b^{th}$ block with local block coordinate
$i\in\{1,\ldots,\dmx\}$ corresponds to global index $b\,\dmx+i-1$;
thus $(\y_b)_i=y_{b\dmx+i-1}$.
Each scale
remains fixed within its block, while the trellis state and accumulated
cost continue across block boundaries.

\input{section_files/mx_hash}

\subsection{Relaxed Joint Scale Search}
\label{sec:joint-scale-search}

A straightforward approach to encoding $\y$ is to first choose the scale $s_b$ for each MX block according to a heuristic rule, such as scaling the maximum to fit in the element quantizer's range (see \cref{sec:mx-background}).
The normalized blocks are then quantized via beam search, with each $s_b$ held fixed.
To improve quantization accuracy, we \emph{jointly} optimize block scales and trellis encodings during beam search.
Unlike NSNQuant's post-quantization scale adjustment~\citep{son2025nsnquant}, this requires ranking partial paths while scale updates change the reconstruction error of previous values.

We relax the quantized block scale $s_b$ to a continuous variable $a_b$ and refit it for each candidate path after every beam-search step.
To stabilize estimates from short prefixes, we add the penalty $\lambda(a_b-\heuristicscale_b)^2$ to the prefix reconstruction error, where $\lambda\geq0$ and $\heuristicscale_b=Q_\alpha(\mxrefscale_b)$ is a heuristic reference scale computed from the full block before search.
We use the scale-independent clipped MX-Hash variant (Appendix~\ref{apx:clipped-mx-hash}), so each path's quantized outputs remain fixed as $a_b$ varies.
The regularized objective is therefore quadratic in $a_b$, and its minimizer has a closed form using two candidate-specific sufficient statistics: the inner product of the target and quantized outputs, and the squared norm of those outputs, accumulated over the current block prefix.

Updating these statistics at each step lets us refit the scale without revisiting earlier outputs.
We then round the fitted scale using $Q_\alpha$ and rank candidates by the regularized prefix error at that rounded scale plus the accumulated error of completed blocks.
At block completion, we refit without regularization and round to obtain the final block scale.
Appendix~\ref{app:joint-scale-search} gives the derivation. Algorithm~\ref{alg:joint-scale-score} illustrates the candidate scoring procedure.

\begin{table*}[t]
  \centering
  \input{includes/results/main_results}

  \par\medskip
  \begin{minipage}[t]{0.48\textwidth}
    \vspace{0pt}
    \input{includes/results/ablation_components}
  \end{minipage}\hfill
  \begin{minipage}[t]{0.48\textwidth}
    \vspace{0pt}
    \input{includes/results/ablation_bpe_vs_ppl}
  \end{minipage}
\end{table*}

\subsection{Mixed Precision KV Quantization}

Prior work has shown that keys are more sensitive to quantization
than values~\cite{pmlr-v267-li25dd, tao-etal-2025-asymkv}.
Our payload-allocation analysis in
Figure~\ref{fig:kv-transition-bits} supports this observation
for \MN{}, showing that allocating more bits to keys can improve
accuracy under a fixed total payload budget.
Following this approach, we extend \MN{} to support different
transition-bit budgets for the key and value trellis encoders,
assigning more bits per transition to keys and fewer to values.

This allocation becomes more flexible as the vector length
$\qtipwidth$ increases.
Since each trellis step emits $\qtipwidth$ scalar values, a
one-bit change in the transition budget changes the payload by
$1/\qtipwidth$ bits per element.
Larger vectors therefore enable finer-grained allocation of the
available payload between keys and values.

The computed reconstruction mapping of \MN{} supports this
flexibility without requiring an enlarged lookup table.
For an explicit, unstructured vector codebook, representing
$\qtipwidth$ values at $\qtiprate$ bits per element requires
$2^{\qtiprate\qtipwidth}$ entries.
Its storage requirements therefore constrain the vector lengths
and payload allocations that are practical.





%% file: section_files/mx_hash.tex
\subsection{MX-Hash}
\label{sec:mx-hash}

\input{section_files/mx_gaussian}

\paragraph{State Mixing.} Here, we introduce the bit mixing stage that takes a trellis state
$u\in\mathcal U=\{0,\ldots,2^L-1\}$ and mixes it before applying the inverse CDF. 
Each trellis state produces a vector with $V$ output coordinates. The mixer therefore consists of $V$ distinct mixer functions that share the same structure but use different parameterizations. 
Specifically, to produce the $v$th output coordinate, we first apply a Linear Congruential Generator (LCG) with fixed unsigned 32-bit coefficients $A_v$ and $C_v$, given by:
\begin{equation}
\lcg_{\hashparams}(u)
=(A_v u+C_v)\bmod2^{32}.
\label{eq:mx-hash-lcg}
\end{equation}
After the LCG we apply the Mixer function parameterized by $\hashparams_v=\{A_v,C_v,r_{v,1},r_{v,2},r_{v,3},m_{v,1},m_{v,2}\}$ and  is given by:
\begin{equation}
\begin{aligned}
\mxmix_{\hashparams_v}(w)&=
\mxxorshift\!\left(
\mxmixstep\!\left(\mxmixstep(w;r_{v,1},m_{v,1});
r_{v,2},m_{v,2}\right);r_{v,3}\right),\\
w_v&=\mxmix_{\hashparams_v}
\!\left(\lcg_{\hashparams,v}(u)\right),
\end{aligned}
\label{eq:mx-hash-mixer}
\end{equation}
where $\mxmixstep(x;r,m)=(m[x\oplus(x\gg r)])\bmod2^{32}$
combines an XOR with logical right shift $r$ and multiplication by $m$,
$\mxxorshift(x;r)=x\oplus(x\gg r)$ performs the final XOR shift, $\oplus$ denotes bitwise XOR and $\gg$ denotes logical right shift. By aggregating all $V$ mixing functions, we obtain the parameter set $\Omega=\{\Omega_{v}\}_{v=1}^V$, with parameters selected using the MX-Hash calibration procedure described in Appendix~\ref{apx:implementation-details}.


\paragraph{MX-Gaussian CDF hash.}
Using the mixed word $w_\ell$ defined above, extract its sign and magnitude rank:
\begin{equation}
\begin{aligned}
\mxsignbit_\ell&=\left\lfloor\frac{w_\ell}{2^{31}}\right\rfloor,
\qquad U_\ell=w_\ell\bmod2^{31}.
\end{aligned}
\label{eq:mx-hash-sign-rank}
\end{equation}
Thus, $\mxsignbit_\ell$ supplies the sign bit and $U_\ell$ the magnitude rank.
The hash converts the magnitude rank to the interior probability
\begin{equation}
p_\ell=\frac{U_\ell+\tfrac12}{2^{31}}, \qquad 0<p_\ell<1.
\label{eq:mx-hash-interior-probability}
\end{equation}
Let $Q_E$ round to the finite grid $\mxelementgrid$ of the selected element
format, using nearest rounding, ties to even, and finite saturation.
At a represented scale \(s\), use the scale-conditioned inverse
\(F^{-1}(p;s)\) from \cref{thm:mx-gaussian-conditional}.
Using the sign bit and interior probability, the output coordinate mapping and quantized
vector are
\begin{equation}
\begin{aligned}
h_{\hashparams,\ell}(u;s)
&=Q_E\!\left((-1)^{\mxsignbit_\ell}F^{-1}(p_\ell;s)\right),\\
\mxhashvec(u;s)
&=\bigl(h_{\hashparams,1}(u;s),\ldots,
h_{\hashparams,V}(u;s)\bigr)
\in\mxelementgrid^{\,V}.
\end{aligned}
\label{eq:mx-hash-cdf-mapping}
\end{equation}
The floating-point CDF mapping can be implemented using precomputed scale
coefficients and a shared inverse-CDF lookup table, as detailed in Appendix
\ref{app:efficient-lut}.


To compare quantizer parameters at different effective bit rates, we report $\mathrm{RMSE}\cdot 2^{b_{\mathrm{eff}}}$, where $\mathrm{RMSE}$ is the root mean square reconstruction error and $b_{\mathrm{eff}}$ is the effective number of bits per element. This compensates for the $2^{-b}$ scaling of RMSE predicted by high-rate quantization theory \citep{gray1998quantization}.
Figure~\ref{fig:parameter-analysis} reports results for standard Gaussian inputs across three parameters: subvector length $V$, trellis length $T$, and MX block size. The best results occur at $V=4$, which we hypothesize reflects the relationship between the 16-bit state and 4-bit element width. Longer trellises improve performance, with marginal gains beyond $T=1024$. Finally, larger MX blocks offer little benefit for Gaussian data. 
The corresponding bit budget is better allocated elsewhere, such as to transition bits. However, this does not capture the advantages of MX in terms of hardware compatibility and improvements in compute.

%% file: section_files/mx_gaussian.tex
To derive the MX hash, we begin with the same assumption as QTIP: the data follows a Gaussian distribution, namely $\randomvec{Y}\sim\mathcal N(\mathbf0,\matsym I_{\dmx})$.  
Since the trellis encodes the scaled values
$\randomvec X=\randomvec Y/\mxscale(\randomvec Y)$, we construct
a hash function using inverse transform sampling to match the
distribution of $\randomvec X$, which we call the MX-Gaussian distribution.
Assuming that decoded trellis states are approximately uniformly
distributed, we map each state to a uniform value and apply the
inverse cumulative distribution function (CDF) to obtain a
reconstruction value with the target distribution.

Given a MX block scale $s$ we derive the inverse CDF of MX-Gaussian condition on the scale. Specifically, let 
$\mxcellmass(s;d)\triangleq\stderf\!\left(b(s)/\sqrt2\right)^d
-\stderf\!\left(a(s)/\sqrt2\right)^d>0$ be the scale

probability, $a(s),b(s)$ be lower and upper boundary of the scale $s$, and $p^\star(s)=\stderf\!\left(a(s)/\sqrt2\right)
\mxcellmass(s;\dmx-1)/\mxcellmass(s;\dmx)$. Then for $0<p<1$ the inverse CDF of MX-Gaussian condition on the scale is given by:
\begin{align}\label{eq:mx-gaussian-conditional-inverse}
&F^{-1}(p;s)\\
&=\frac{\sqrt2}{s}\stdierf\!\left(
\begin{cases}
\displaystyle
\frac{\mxcellmass(s;\dmx)p}{\mxcellmass(s;\dmx-1)},
&
p\leq p^\star(s)
,\\
\frac{
\mxcellmass(s;\dmx)p
{}+\stderf\left(a(s)/\sqrt2\right)^{\dmx}
}{\stderf\left(b(s)/\sqrt2\right)^{\dmx-1}},
&
p>p^\star(s).
\end{cases}
\right),\nonumber
\end{align}
where $\stderf$ denote the standard error function and $\stdierf$ its inverse.
The derivation appear in Appendix ~\ref{app:mx-gaussian-proof}. Using \eqref{eq:mx-gaussian-conditional-inverse}, we can sample from the MX-Gaussian distribution given a uniform distribution. In the following paragraph, we first introduce bit mixing to improve our ability to generate a uniform distribution. Then, in the next paragraph, we combine this with the inverse CDF to produce the MX-Gaussian hash.

%% file: includes/results/main_results.tex
\centering
\color{black}
\caption{WikiText-2 perplexity, mean LongBench score, and GSM8K accuracy (\%) for MX KV-cache quantization. 
Effective bits include payload and stored quantization metadata. 
Blue marks the best quantized result, including ties, per model and metric within each rate and format group.} 
\label{tab:main-table}
\begingroup
\footnotesize
\setlength{\tabcolsep}{3pt}
\setlength{\arrayrulewidth}{0.3pt}
\renewcommand{\arraystretch}{1.35}
\resizebox{\textwidth}{!}{%
\begin{tabular}{@{}clc|ccc|ccc@{}}
\hline
\multirow{2}{*}{\shortstack{Block-scale\\format}} & \multirow{2}{*}{Variant} & \multirow{2}{*}{\shortstack{Effective\\Bit}} & \multicolumn{3}{c|}{\textbf{Llama-3.1-8B-Instruct}} & \multicolumn{3}{c}{\textbf{Qwen3-8B}} \\
& & & Perplexity $\downarrow$ & LongBench $\uparrow$ & GSM8K $\uparrow$ & Perplexity $\downarrow$ & LongBench $\uparrow$ & GSM8K $\uparrow$ \\
\hline
--- & BF16 & 16.00 & 7.170 & 48.50 & 76.95 & 9.750 & 41.07 & 88.63 \\
\hline
\multirow{3}{*}{UE8M0} & VecInfer~\cite{yao2026vecinfer} & 2.25 & 10.985 & 42.24 & 23.12 & 10.687 & 38.15 & 68.76 \\
 & NSNQuant~\cite{son2025nsnquant} & 2.24 & 7.806 &  47.19 & 64.97 & 10.386 &  40.39 & 80.67 \\
 & \TableMN{}  & 2.25 & \cellcolor[HTML]{9DC3E6}\color{black} 7.743 & \cellcolor[HTML]{9DC3E6}\color{black} 47.60 & \cellcolor[HTML]{9DC3E6}\color{black} 69.60 & \cellcolor[HTML]{9DC3E6}\color{black} 10.217 & \cellcolor[HTML]{9DC3E6}\color{black} 40.46 & \cellcolor[HTML]{9DC3E6}\color{black} 83.62 \\
\hline
\multirow{3}{*}{UE5M3} & VecInfer~\cite{yao2026vecinfer} & 2.50 & 8.022 & 43.58 & 54.13 & 10.275 & 38.62 & 75.74 \\
 & NSNQuant~\cite{son2025nsnquant} & 2.24 & 7.808 & 47.62 & 67.17 & 10.370 & \cellcolor[HTML]{9DC3E6}\color{black} 41.16 & \cellcolor[HTML]{9DC3E6}\color{black} 81.50 \\
 & \TableMN{}  & 2.25 & \cellcolor[HTML]{9DC3E6}\color{black} 7.642 & \cellcolor[HTML]{9DC3E6}\color{black} 48.35 & \cellcolor[HTML]{9DC3E6}\color{black} 70.13 & \cellcolor[HTML]{9DC3E6}\color{black} 10.173 & 40.62 & 75.97 \\
\hline
\multirow{3}{*}{NVFP4} & VecInfer~\cite{yao2026vecinfer} & 2.25 & 10.782 & 42.70 & 27.00 & 10.526 & 39.60 & 70.58 \\
 & NSNQuant~\cite{son2025nsnquant} & 2.24 & \cellcolor[HTML]{9DC3E6}\color{black} 7.769 & \cellcolor[HTML]{9DC3E6}\color{black} 47.89 &  64.59 & 10.395 & 40.00 &  80.59 \\
 & \TableMN{}  & 2.25 & 7.879 & 47.80 & \cellcolor[HTML]{9DC3E6}\color{black} 67.70 & \cellcolor[HTML]{9DC3E6}\color{black} 10.343 & \cellcolor[HTML]{9DC3E6}\color{black} 40.63 & \cellcolor[HTML]{9DC3E6}\color{black} 83.02 \\
\hline
\hline
\multirow{3}{*}{UE8M0} & VecInfer~\cite{yao2026vecinfer} & 1.25 & 258.713 & 1.20 & 1.97 & 35.042 & 6.48 & 2.20 \\
 & NSNQuant~\cite{son2025nsnquant} & 1.24 & 15.029 & \cellcolor[HTML]{9DC3E6}\color{black} 38.23 & 22.90 & 20.457 &  30.96 & 27.14 \\
 & \TableMN{}  & 1.25 & \cellcolor[HTML]{9DC3E6}\color{black} 11.957 & 35.79 & \cellcolor[HTML]{9DC3E6}\color{black} 33.06 & \cellcolor[HTML]{9DC3E6}\color{black} 15.072 & \cellcolor[HTML]{9DC3E6}\color{black} 31.71 & \cellcolor[HTML]{9DC3E6}\color{black} 45.94 \\
\hline
\multirow{3}{*}{UE5M3} & VecInfer~\cite{yao2026vecinfer} & 1.25 & 232.335 & 1.22 & 1.90 & 30.791 & 11.43 & 2.50 \\
& NSNQuant~\cite{son2025nsnquant} & 1.24 & 14.968 & 38.15 & 23.12 & 20.319 &  32.16 & 29.04 \\
 & \TableMN{}  & 1.25 & \cellcolor[HTML]{9DC3E6}\color{black} 11.630 & \cellcolor[HTML]{9DC3E6}\color{black} 40.37 & \cellcolor[HTML]{9DC3E6}\color{black} 35.48 & \cellcolor[HTML]{9DC3E6}\color{black} 14.657 & \cellcolor[HTML]{9DC3E6}\color{black} 33.42 & \cellcolor[HTML]{9DC3E6}\color{black} 52.46 \\
\hline
\multirow{3}{*}{NVFP4} & VecInfer~\cite{yao2026vecinfer} & 1.25 & 234.929 & 1.64 & 1.59 & 30.585 & 10.21 & 2.65 \\
 & NSNQuant~\cite{son2025nsnquant} & 1.24 & 14.787 & 37.92 & 23.65 & 20.244 & 31.34 & 29.57 \\
 & \TableMN{}  & 1.25 & \cellcolor[HTML]{9DC3E6}\color{black} 11.469 & \cellcolor[HTML]{9DC3E6}\color{black} 41.02 & \cellcolor[HTML]{9DC3E6}\color{black} 37.23 & \cellcolor[HTML]{9DC3E6}\color{black} 14.362 & \cellcolor[HTML]{9DC3E6}\color{black} 33.06 & \cellcolor[HTML]{9DC3E6}\color{black} 50.11 \\
\hline
\end{tabular}%
}
\endgroup

%% file: includes/results/ablation_components.tex
\centering
\color{black}
\caption{WikiText-2 perplexity ($\downarrow$) across quantizer
designs at approximately 1.30 effective bits per element with a UE5M3 scale format,
providing a comparison at similar storage budgets.
Blue cells mark the lowest perplexity for each model.}
\label{tab:components-abl}
\begingroup
\footnotesize
\setlength{\tabcolsep}{3pt}
\setlength{\arrayrulewidth}{0.3pt}
\renewcommand{\arraystretch}{1.35}
\resizebox{\linewidth}{!}{%
\begin{tabular}{@{}l|c|c@{}}
\hline
Variant & \textbf{Llama-3.1-8B-Instruct} & \textbf{Qwen3-8B} \\
\hline
NSN VQ MX & 12.966 & 16.530 \\
Naive MX Trellis & 11.530 & 15.682 \\
MX Trellis Hash & 11.195 & 15.431 \\
MX Trellis Hash Joint & \cellcolor[HTML]{9DC3E6}\color{black} 10.699 & \cellcolor[HTML]{9DC3E6}\color{black} 14.942 \\
\hline
\end{tabular}
}
\endgroup


%% file: includes/results/ablation_bpe_vs_ppl.tex
  \centering
\captionsetup{skip=3pt}
    \includegraphics[width=0.85\linewidth]{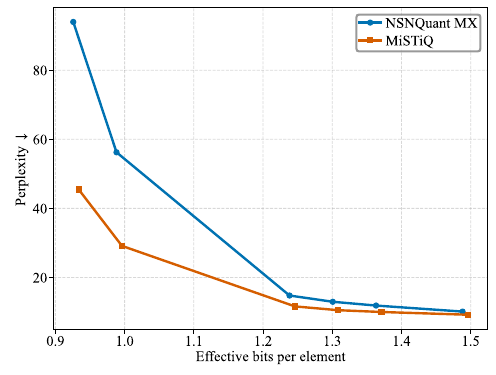}
  \captionof{figure}{WikiText-2 perplexity comparison for decreasing bit rates between \MN{} and NSNQuant~\cite{son2025nsnquant}.}
  \label{fig:bpe-ppl}

%% file: files/theortical_analysis.tex

%% file: files/experiments.tex
\section{Experiments}
\label{sec:experiments}

We evaluate \MN{} on Llama-3.1-8B-Instruct~\cite{dubey2024llama} and Qwen3-8B~\cite{yang2025qwen3} to assess its effectiveness for KV-cache quantization at very low bit widths. 
We compare \MN{} with existing vector quantization methods designed for KV-cache compression, evaluating WikiText-2 perplexity (PPL)~\cite{merity2016pointer}, performance across LongBench tasks~\cite{bai-etal-2024-longbench}, and accuracy on GSM8K~\cite{cobbe2021gsm8k} under a decode-heavy chain-of-thought generation setting. 
We further conduct ablation studies to isolate the contributions of individual components and assess the trade-offs underlying our design choices. 
We present sub-1-bit PPL results in Appendix~\ref{apx:below_1bit_res}.

\begin{figure*}[t]
  \centering
  \begin{minipage}[t]{0.48\textwidth}
    \vspace{0pt}
    \input{includes/results/ablation_vector_length}
  \end{minipage}\hfill
  \begin{minipage}[t]{0.48\textwidth}
    \vspace{0pt}
    \input{includes/results/ablation_kv_payload}
  \end{minipage}
\end{figure*}

\subsection{Experimental Settings}
For the main results, we configure the \MN{} quantizer with a block size of 32 (16 for NVFP4) at effective bit widths above 2 bits per element, and 128\footnote{Note that hardware with a small block size (e.g., 16 or 32) can accommodate larger block sizes by replicating the scale.} below 2 bits per element. These choices are intended to maintain compatibility with MX-capable hardware while allocating more of the storage budget to trellis-coded values.
For the UE5M3 and NVFP4 configurations, we use the relaxed joint scale search described in Section~\ref{sec:joint-scale-search}.


For the main results, we integrate \MN{} into the NSNQuant~\cite{son2025nsnquant} workflow as a drop-in replacement for its codebook-based vector quantizer, retaining its normalization and Hadamard transformations.
Normalization is applied before RoPE, and \MN{} quantization is applied to the KV cache after RoPE.
Appendix~\ref{apx:vec-comparison} demonstrates
this integration with VecInfer, improving its accuracy below 2 effective bits per element.

Unless otherwise stated, the trellis encoder in \MN{} uses a state-bit length of $\qtipmemory=16$ and a source vector length of $T=1024$, with $\qtipwidth=4$ scalar values emitted per trellis step in the higher-bit settings and $\qtipwidth=8$ in the lower-bit settings. Encoding uses beam search with beam width $B=128$.
We implement the baseline methods following their original descriptions and adapt them to MX formats by quantizing their codebook elements to the target MX element format.
All experiments were conducted on NVIDIA H100 GPUs, using one GPU per run except for GSM8K evaluations, which used four GPUs.
Full implementation details, including per-variant transition-bit allocation, are provided in Appendix~\ref{apx:implementation-details}.

\subsection{Accuracy Evaluation}

Table~\ref{tab:main-table} compares \MN{} with the baseline KV-cache vector quantization methods across a range of evaluation workloads, from WikiText-2 perplexity evaluation without free-form generation to long-context understanding and decode-heavy generation, including summarization and GSM8K reasoning. 
For LongBench, we evaluate seven tasks covering question answering, summarization, and classification: Qasper~\cite{dasigi-etal-2021-dataset}, TriviaQA~\cite{joshi-etal-2017-triviaqa}, GovReport~\cite{huang-etal-2021-efficient-gov}, QMSum!\cite{zhong-etal-2021-qmsum}, MultiNews~\cite{fabbri-etal-2019-multi}, SAMSum~\cite{gliwa-etal-2019-samsum}, and TREC~\cite{li-roth-2002-learning-trec}. 
We report the unweighted average of their scores in the main table, with full per-task results provided in Appendix~\ref{apx:longbench_full}.

The results show that \MN{} achieves strong overall performance, with particularly pronounced gains in the lower-bit regime.
At approximately $1.25$ effective bits per element, it achieves the lowest perplexity and highest GSM8K accuracy among the compared quantized methods on both models across all three block-scale configurations.
With UE5M3 at this rate, it also achieves the highest LongBench average on both models.

We further extend the comparison with NSNQuant to decreasing payload budgets in Figure~\ref{fig:bpe-ppl}.
Here, \MN{} uses a vector length of $\qtipwidth=4$, while NSNQuant varies the key and value codebook sizes to obtain different storage budgets.
\MN{} consistently achieves lower perplexity at comparable effective bit widths, with an increasingly pronounced advantage under extreme compression, particularly below one effective bit per element.


\subsection{Ablation Study}

\subsubsection{Quantizer Design Comparison}

Table~\ref{tab:components-abl}  evaluates the successive design choices underlying \MN{} using WikiText-2 perplexity at approximately matched effective storage budgets.
Replacing the NSNQuant-style codebook quantizer with trellis quantization followed by MX conversion (\textbf{Naive MX Trellis}) substantially reduces perplexity on both models.
Incorporating MX element constraints directly into the reconstruction mapping (\textbf{MX Trellis Hash}) yields further improvements by making the trellis search aware of the final reconstruction values.
Jointly optimizing the trellis path and block scales (\textbf{MX Trellis Hash Joint}) further reduces perplexity, achieving the best results on both models.
Together, these results show that trellis quantization provides the largest initial improvement, while MX-aware reconstruction and joint scale search contribute additional gains.


\subsubsection{Vector Length and Payload Allocation}

We next examine how the number of scalar values emitted per trellis step, $\qtipwidth$ (referred to as the ``vector length''), affects quantization accuracy and the flexibility of payload allocation between keys and values. 
We first evaluate vector length under uniform payloads in Figure~\ref{fig:qtip-vector-length}, then show how the finer-grained allocation enabled by larger vectors can further improve accuracy with mixed precision payloads in Figure~\ref{fig:kv-transition-bits}.

Although a vector length of $\qtipwidth=4$ yields the lowest perplexity on both models under uniform payloads, $\qtipwidth=8$ enables finer-grained payload allocation. 
This flexibility allows mixed precision configurations to assign a larger payload to the more sensitive keys, further reducing perplexity.
This flexibility is particularly valuable for extremely low effective bit regimes, where small changes in payload allocation can substantially affect accuracy, motivating our use of $\qtipwidth=8$ in this regime and $\qtipwidth=4$ in the higher-bit configurations.

%% file: includes/results/ablation_vector_length.tex
\centering
\captionsetup{skip=3pt}
    \includegraphics[width=0.85\linewidth]{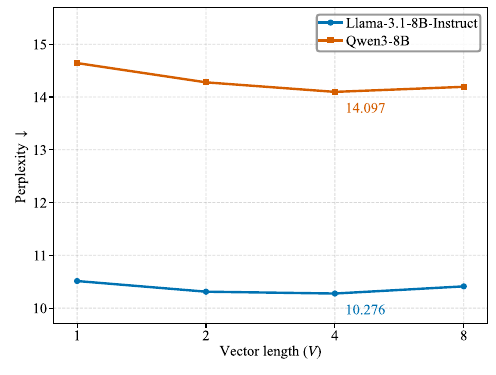}
\captionof{figure}{WikiText-2 perplexity versus vector length for \MN{} with UE5M3 scales, $\dmx=128$, and $T=128$ and a trellis payload of 1 bit per element.}
\label{fig:qtip-vector-length}

%% file: includes/results/ablation_kv_payload.tex
\centering
\captionsetup{skip=3pt}
    \includegraphics[width=0.85\linewidth]{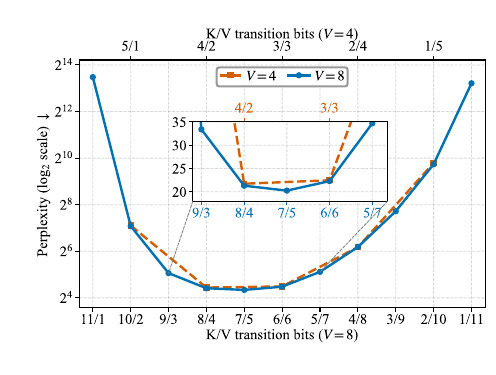}
\captionof{figure}{WikiText-2 perplexity versus K/V payload allocation for Llama-3.1-8B-Instruct, with UE5M3 scales, $\dmx=128$, and $T=1024$. Upper and lower x-axis labels show K/V bits per transition for $\qtipwidth=4$ and $\qtipwidth=8$, respectively.}
\label{fig:kv-transition-bits}

%% file: files/conclusion.tex
\section{Conclusions}

In this work, we presented \MN{}, a KV-cache quantizer that combines compact trellis-coded storage with microscaling formats for computation, bringing MX quantization to the KV cache.
We examined how to preserve accuracy at very low storage budgets while reconstructing the cache directly in MX-compatible representations.
To this end, we developed a beam-search trellis encoder that limits the retained search space, enabling KV-cache quantization with large reconstruction spaces without explicitly storing large vector codebooks.
We further introduced MX-Hash functions that generate reconstruction values directly in the target MX element format.
We also developed relaxed joint search over trellis encodings and block scales, incorporating MX constraints into the encoding process.
Experiments across language modeling, long-context understanding, and reasoning demonstrate strong overall performance, with particularly substantial gains at low effective bit widths.

These results motivate extending the combination of trellis-coded storage and MX computation to weight quantization, where offline encoding permits more extensive search.
Supporting a broader range of weight and KV-cache distributions also motivates relaxing the Gaussian assumption underlying our hash construction, for example through reconstruction mappings adapted to empirical distributions.

%% file: section_files/implementation_details.tex
\section{Implementation Details}
\label{apx:implementation-details}

\paragraph{Quantizer implementation.}
\MN{} encodes vectors as trellis paths and represented block scales,
using Triton kernels with FP32 arithmetic during search.
The main implementation uses $\qtipmemory=16$ state bits,
$\beamstartbits=8$ initial-state selection bits, $T=1024$, and beam width $B=128$,
unless otherwise specified. Reconstruction elements lie directly
on the E2M1 grid, avoiding a second MX conversion after encoding.
The quantizer can be integrated with different preprocessing
workflows. We apply key-channel standardization by subtracting
a frozen per-channel mean and dividing by a frozen per-channel
standard deviation before the Hadamard transform. These statistics
are calibrated separately for each layer, KV head, and key channel
in the corresponding preprocessing domain, with standard deviations
floored at $10^{-6}$. After reconstruction and inverse Hadamard
transformation, we restore the original channel scales and means.

\paragraph{MX-Hash calibration.}
Hash resources are prepared separately for each element format, scale format, vector length, and search policy. For the initial-scale variant, integer CDF boundaries are precomputed for each represented scale value, allowing a 31-bit hash rank to map directly to an MX element without evaluating an inverse CDF at runtime. The mixer parameters for each output coordinate are calibrated jointly over synthetic inputs, rather than independently for every scale value. The joint-scale variant instead calibrates the mixers and the spread of a symmetric Gaussian emission distribution on the native element grid, producing a scale-independent mapping.

\label{app:mixing-parameter}
We select the post-LCG mixing parameters through reproducible
offline random search, keeping the LCG coefficients
$A_\ell,C_\ell$ fixed.
Each candidate jointly specifies all $\qtipwidth$ output coordinates:
three shifts per output coordinate sampled uniformly from $\{1,\ldots,31\}$
and two odd unsigned 32-bit multipliers sampled as $2q+1$,
where $q$ is uniform on $\{0,\ldots,2^{31}-1\}$.
All candidates encode the same fixed Gaussian training vectors
under identical search settings. We minimize the mean
reconstruction error in the original domain:
\[
J(\Omega)=\frac{1}{NT}\sum_{n=1}^{N}
\left\|\mathbf y^{(n)}
-\widehat{\mathbf y}^{(n)}(\Omega)\right\|_2^2,
\]
where $N$ is the number of training vectors, $T$ is their
length, and $\Omega$ contains the mixing parameters for all
output coordinates. Reconstruction includes the represented block scales
and element-format rounding. Ties are resolved by candidate
order, and the selected parameters are frozen for evaluation
and decoding.

The standard calibration evaluates 48 candidates using
32 training vectors of length 128 and 256 held-out vectors.
The initial E2M1 mixers retain their original scale-node calibration objective, while deployment fixes the scale before search. Joint resources use the fitted-and-rounded scale objective with $\alpha=1$. The higher-bit NVFP4 resource is calibrated separately using eight training and 32 held-out vectors of length 1024, minimizing the mean reconstruction error for its K/V payloads. Larger runtime blocks reuse the recorded calibrated mappings without recalibration.

\paragraph{Block-scale computation and search.}
For each input block, we first compute the reference scale as its maximum absolute value divided by the element-format maximum, which is 6 for E2M1, and round it to the selected scale format. In the \emph{initial-scale} option, this represented scale remains fixed throughout encoding, and the hash uses its scale-conditioned mapping. In the \emph{joint-scale} option, each retained path accumulates the target--emission inner product and emission squared norm. These statistics support a continuous least-squares scale update at each step, regularized toward the reference scale for incomplete blocks. The fitted scale is rounded before candidate scoring; at block completion, the scale is refitted without regularization. The joint mapping remains independent of the fitted scale, so decoding requires only one stored output scale per block.

Our experiments use initial-scale search with UE8M0 and joint-scale search with UE5M3 and NVFP4. UE8M0 scales round upward; UE5M3 uses nearest rounding with ties toward zero. NVFP4 uses E4M3FN block scales with nearest-even rounding and an additional per-tensor global scale. Its lower-bit block-size-128 configuration transfers a block-size-16 calibrated mapping and is therefore identified as NVFP-style.

\paragraph{Mixed-bit payload implementation.}
Keys and values are encoded separately with independently specified integer transition-bit budgets, while retaining the same vector length and reconstruction format. A budget of $c$ bits gives $2^c$ successors per trellis state and a payload of $c/\qtipwidth$ bits per scalar element. Increasing $\qtipwidth$ therefore permits finer payload increments without changing the MX element precision. The allocation is fixed for each role across layers; it does not introduce token-dependent or layer-dependent bit selection.

\paragraph{Evaluated payload configurations.}
The higher-bit UE8M0/UE5M3 configurations use $\qtipwidth=4$, block size 32, and K/V transition budgets of 8/6. Higher-bit NVFP4 uses $\qtipwidth=4$, block size 16, and budgets of 7/5. Lower-bit configurations use $\qtipwidth=8$, block size 128, and budgets of 9/6. The uniform-payload vector-length ablation fixes block size 128, $T=128$, and one payload bit per element for both roles. The tighter payload-allocation sweep compares $\qtipwidth=4$ and 8 at block size 128 and $T=1024$, keeping the sum of the K/V payloads fixed at 1.5 bits per element. These ablations use UE5M3 joint-scale search with $B=128$ and the WikiText-2 perplexity protocol described in the experimental setup.

\paragraph{Baseline methods with MX compute formats.}
We adapt NSNQuant and VecInfer by converting their reconstructed VQ payloads to the target MX format before inverse transformation and metadata restoration. Their codebook fitting and centroid assignment retain FP32 codebooks and do not incorporate the output MX conversion into their objectives. NSNQuant uses a signed, 256-entry Gaussian codebook shared across layers and K/V roles, trained with 30 Lloyd iterations. It retains both normalization stages, 64-token windows, and full-head runtime Hadamard transformations; its gain is fitted from the raw VQ reconstruction before MX conversion. VecInfer uses reciprocal post-RoPE query/key smoothing, full-head query/key Hadamard transformations, and unrotated values. Its layer- and role-specific codebooks use 1024/256 K/V entries at vector length 4 and 4096/256 entries at vector length 8, with 30 fitting iterations and Triton assignment. Both methods recompute output-MX scales during decoding, so these scales add no retained-cache bits. Output conversion uses block size 32 for UE8M0/UE5M3 and 16 for the NSNQuant NVFP4 baseline.

%% file: section_files/alg_pseudo_code.tex
\section{Algorithm Pseudo Code}
\label{app:detailed-implementation}

\subsection{Backpointers and traceback}
\label{app:beam-backpointers}

For each selected expansion $(\beamhyp,v)$, scored using the
relaxed joint scale objective in Section~\ref{sec:joint-scale-search},
the encoder assigns a new hypothesis index $\beamhyp'$ and stores
\begin{equation}
\begin{aligned}
\beamterminal_t(\beamhyp')&=v,\\
\beamcost_t(\beamhyp')&=\widetilde{\beamcost}_t(\beamhyp,v),\\
\beambackpointer_t(\beamhyp')&=\beamhyp.
\end{aligned}
\label{eq:beam-backpointer}
\end{equation}
The backpointer identifies the predecessor hypothesis at step $t-1$,
including its own cost and path history, even when several hypotheses
share a terminal state. Traceback recovers the selected initial state, trellis path,
and block scales. Backpointers are not stored in the encoded cache.
Enumerate predecessor hypotheses and their branches in a fixed order;
equal-cost candidates retain this enumeration order during top-$B$
selection. Terminal ties use the retained hypothesis order.
Hypotheses with the same terminal state occupy separate beam slots.
Because every QTIP state has nonempty successors and $B\geq1$,
$\beamset_n$ is nonempty.

\begin{algorithm}[H]
\caption{Trellis encoding with joint scale search}
\label{alg:trellis_beam_encoding}
\begin{algorithmic}[1]
\Require \(n\geq1\), \(\trellissource_{1:n}\), QTIP bitshift trellis
         \(\graphG=(\mathcal U,\mathcal A)\),
         scale-independent MX-Hash map \(C_{\mathrm{MX}}\),
         reference block scales \(\{s_b^{\mathrm{ref}}\}\),
         block size \(\dmx\), vector length \(V\),
         regularization weight \(\lambda\geq0\), integer \(B\geq1\),
         integer \(0\leq\beamstartbits\leq L\)
\Ensure Initial state \(u_0\), state sequence \(u_{1:n}\),
        and UE5M3 block scales \(\{s_b\}\)
\Statex \textbf{Tie rule:} fixed hypothesis and branch enumeration order.
\State \(\beamset_0\gets\{0,\ldots,2^{\beamstartbits}-1\}\)
\State For each \(\beamhyp\in\beamset_0\), set
       \(\beamterminal_0(\beamhyp)\gets\beamhyp\),
       \(\beamcost_0(\beamhyp)\gets0\), and
       \(\mathbf{S}_\beamhyp\gets(0,0,0)\)
\For{\(t=1,\ldots,n\)}
    \State \(\beamcandidates_t\gets\) empty candidate list
    \ForAll{\(\beamhyp\in\beamset_{t-1}\)}
        \ForAll{\(v\in\operatorname{succ}(\beamterminal_{t-1}(\beamhyp))\)}
            \State \(\widetilde{\beamcost}_t(\beamhyp,v)
            \gets\operatorname{JointScaleScore}_t(\mathbf{S}_\beamhyp,v;\lambda)\)
            \State Append candidate
                   \((\beamhyp,v,\widetilde{\beamcost}_t(\beamhyp,v))\)
                   to \(\beamcandidates_t\)
        \EndFor
    \EndFor
    \State Retain the \(\min(B,|\beamcandidates_t|)\) lowest-cost
       candidates across all parents and index them by \(\beamset_t\)
    \ForAll{selected candidates \((\beamhyp,v,\widetilde{\beamcost}_t(\beamhyp,v))\) with new index \(\beamhyp'\)}
        \State \(\beamterminal_t(\beamhyp')\gets v\);
               \(\beamcost_t(\beamhyp')\gets\widetilde{\beamcost}_t(\beamhyp,v)\);
               \(\beambackpointer_t(\beamhyp')\gets\beamhyp\)
    \EndFor
\EndFor
\State \(\beamhyp\gets\operatorname*{arg\,min}_{\beamhyp'\in\beamset_n}\beamcost_n(\beamhyp')\)
\For{\(t=n,n-1,\ldots,1\)}
    \State \(u_t\gets\beamterminal_t(\beamhyp)\)
    \If{\(tV\bmod\dmx=0\)}
        \State \(s_{tV/\dmx-1}\gets\sigma_t(\beamhyp)\)
    \EndIf
    \State \(\beamhyp\gets\beambackpointer_t(\beamhyp)\)
\EndFor
\State \(u_0\gets\beamterminal_0(\beamhyp)\)
\State \Return \((u_0,u_{1:n},\{s_b\})\)
\end{algorithmic}
\end{algorithm}

Algorithm~\ref{alg:joint-scale-score} details the candidate scoring procedure.
The target blocks, reference scales, emission map, and scale quantizer
are shared with Algorithm~\ref{alg:trellis_beam_encoding}.
Candidates with nonpositive fitted scales are discarded.

\begin{algorithm}[H]
\caption{JointScaleScore$_t(\mathbf{S}_\beamhyp,v;\lambda)$}
\label{alg:joint-scale-score}
\begin{algorithmic}[1]
\Require Parent statistics \(\mathbf{S}_\beamhyp=(R_\beamhyp,W_\beamhyp,D_\beamhyp)\),
         successor \(v\), regularization weight \(\lambda\geq0\)
\Ensure Candidate score, updated tuple \(\mathbf{S}'\), and completed-block scale \(\sigma_t\)

\State \(b\gets\lfloor(t-1)V/\dmx\rfloor\);
       \(m\gets1+((t-1)\bmod(\dmx/V))\)
\State \(\lambda'\gets\lambda\text{ if }tV\bmod\dmx\ne0\text{ else }0\)
       \Comment{Block completion: disable regularization}
\State \(\widehat{\x}\gets C_{\mathrm{MX}}(v)\)
\State \(R\gets R_\beamhyp+\|\widehat{\x}\|_2^2\)
\State \(W\gets W_\beamhyp+
       \langle\y_{b,m},\widehat{\x}\rangle\)
\State \(a\gets(W+\lambda'\heuristicscale_b)/(R+\lambda')
       \enspace\text{if }R>0\text{ else }\heuristicscale_b\)
       \Comment{Fit block scale}

\State \(s\gets Q_\alpha(a)\)
\State \(J\gets D_\beamhyp+s^2R-2sW
       +\lambda'(s-\heuristicscale_b)^2\)

\If{\(tV\bmod\dmx=0\)}
    \State Attach \(\mathbf{S}'=(0,0,J)\) and block scale \(\sigma_t=s\) to the candidate
           \Comment{Block completion: reset statistics}
\Else
    \State Attach \(\mathbf{S}'=(R,W,D_\beamhyp)\) to the candidate
\EndIf
\State \Return \(J\)
\end{algorithmic}
\end{algorithm}

\begin{algorithm}[H]
\caption{Compact trellis decoding}
\label{alg:trellis_direct_decoding}
\begin{algorithmic}[1]
\Require Branch labels \(c_{1:n}\), stored initial state \(u_0\),
         block scales \(\{s_b\}\), \(L\), \(k\), \(V\), \(\dmx\),
         scale-independent MX-Hash map \(C_{\mathrm{MX}}\)
\Statex Each \(c_t\in\{0,\ldots,2^{kV}-1\}\) occupies \(kV\) bits.
\Statex The initial state \(u_0\) occupies \(\beamstartbits\) bits; when
        \(\beamstartbits=0\), it is implicitly zero and is omitted.
\Ensure Reconstructed sequence \(\trellisrecon\in\mathbb{R}^{T}\)
\For{\(t=1,\ldots,n\)}
    \State \(u_t\gets(u_{t-1}2^{kV}\bmod 2^L)+c_t\)
    \State \(b\gets\lfloor (t-1)V/\dmx\rfloor\)
    \State \(\trellisrecon_t\gets s_b\,C_{\mathrm{MX}}(u_t)\)
\EndFor
\State \(\trellisrecon\gets
       \operatorname{concat}(\trellisrecon_1,\ldots,\trellisrecon_n)
       \in\mathbb{R}^{T}\)
\State \Return \(\trellisrecon\)
\end{algorithmic}
\end{algorithm}

\paragraph{Storage and candidate handling.}
Retained costs require $O(2^L)$ memory for Viterbi. Beam search requires
$O(2^{\beamstartbits})$ cost storage at initialization and $O(B)$ after pruning.
Stored backpointers require $O((T/V)2^L)$ memory for Viterbi and $O((T/V)B)$
for beam search. If all candidates are materialized, beam search additionally
requires $O(2^{\beamstartbits+kV})$ temporary storage at the first step and
$O(B2^{kV})$ thereafter, although candidates can instead be scored and
selected in a streaming fashion.

\paragraph{Efficient implementation.}
An efficient implementation evaluates candidate transitions in
parallel~\citep{chen2018gpuwfst}, keeps costs and backpointers on the
accelerator, fuses candidate scoring with top-\(B\) selection where practical,
reuses fixed buffers~\citep{wang2021lightseq}, avoids host--device
synchronization inside the recurrence, and batches independent trellises to
improve hardware utilization.

%% file: files/proof.tex
\section{MX-Gaussian CDF and Inverse CDF}
\label{sec:proof}

%% file: section_files/mx_gaussian_proof.tex
\label{app:mx-gaussian-proof}
In this section, we derive the inverse CDF of the MX-Gaussian distribution. We begin by giving a formal definition of the MX-Gaussian distribution. Based on this definition, we then present its basic properties. Finally, we provide the CDF and its inverse.

\begin{definition}[MX-Gaussian] Let $\randomvec{Y}\sim\mathcal N(\mathbf 0,\matsym{I}_{\dmx})$, where $\dmx\geq2$. Define the random block scale $S=\mxscale(\randomvec{Y})$, and let $s\in\mathcal S$ denote a scale value. Let
\(M=\max_{1\leq i\leq\dmx}|Y_i|\). Assume $\mxscale$ takes values in a countable set $\mathcal S$
and satisfies the following conditions:
(i) $S>0$ almost surely;
(ii) $\Pr(S=s)>0$ for every $s\in\mathcal S$; and
(iii) for every $s\in\mathcal S$, there exist bounds
\(0\leq a(s)<b(s)\leq\infty\) such that
    \begin{equation}
\{\mxscale(\randomvec Y)=s\}=\{a(s)\leq M<b(s)\}.
\label{eq:mx-scale-cell}
\end{equation}
Define $\randomvec{X}=\randomvec{Y}/S$. We call the distribution of $\randomvec{X}$ the MX-Gaussian distribution, denoted by $\mathcal{N_{MX}}(0,\matsym{I}_{\dmx},\dmx)$, where $\matsym{I}_{\dmx}$ denotes the covariance of the underlying Gaussian vector.
    
\end{definition}
As a first step, we derive the block-scale probability lemma.
\begin{lemma}[Block Scale Probability]\label{lemma:block_scale}
    Let $\randomvec{X}\sim \mathcal{N_{MX}}(0,\matsym{I}_{\dmx},\dmx)$ be MX Gaussian random vector. Then, the probability of the scale $s$ is given by:
    \begin{equation}\label{eq:scale_prob}
       \mxcellmass(s;\dmx)=\Pr(\mxscale(\randomvec Y)=s)=\stderf\!\left(b(s)/\sqrt2\right)^{\dmx}-\stderf\!\left(a(s)/\sqrt2\right)^{\dmx},
    \end{equation}
    where $\stderf$ is the standard error function. For each positive integer \(d\), define \(\mxcellmass(s;d)=\stderf(b(s)/\sqrt2)^d-\stderf(a(s)/\sqrt2)^d\).
\end{lemma}
\begin{proof}
By independence of the Gaussian coordinates, \(\Pr(M<c)=\stderf(c/\sqrt2)^{\dmx}\) for \(c\geq0\).
    \begin{equation}
\Pr(\mxscale(\randomvec Y)=s)
=\Pr(M<b(s))-\Pr(M<a(s))
=\stderf\!\left(b(s)/\sqrt2\right)^{\dmx}-\stderf\!\left(a(s)/\sqrt2\right)^{\dmx}
.
\end{equation}
\end{proof}


Using the definition of MX-Gaussian and Lemma~\ref{lemma:block_scale}, we can derive the CDF and the inverse CDF in the following theorem:

\begin{theorem}[MX-Gaussian Magnitude]\label{thm:mx-gaussian-conditional}
Let
\(F(\tau;s)=\Pr(|X_i|\leq \tau\mid\mxscale(\randomvec Y)=s)\quad \forall i\in\{1,\ldots,\dmx\}\)
be the magnitude CDF of MX-Gaussian. Then, for $\tau>0$ we have:
    \begin{equation}
F(\tau;s)=
\frac{
\stderf\!\left(\min\{s \tau,b(s)\}/\sqrt2\right)\stderf\!\left(b(s)/\sqrt2\right)^{\dmx-1}
-\stderf\!\left(\min\{s \tau,a(s)\}/\sqrt2\right)\stderf\!\left(a(s)/\sqrt2\right)^{\dmx-1}
}{\mxcellmass(s;\dmx)},
\label{eq:mx-gaussian-conditional-cdf}
\end{equation}
where $\stderf$ is the standard error function and $\stdierf$ its inverse. In particular, \(F(\tau;s)=1\) for
\(\tau\geq b(s)/s\) and \(F(\tau;s)=0\) for $\tau\leq0$.   

Define
\(p^\star(s)=
\stderf\!\left(a(s)/\sqrt2\right)\mxcellmass(s;\dmx-1)/\mxcellmass(s;\dmx)\). Then for $0<p<1$ the inverse CDF is given by \eqref{eq:mx-gaussian-conditional-inverse}.
\end{theorem}


\begin{proof}[Proof of \cref{thm:mx-gaussian-conditional}]

For each coordinate \(j\in\{1,\ldots,\dmx\}\), let \(Z_j=|Y_j|\). For \(\tau\geq0\), set \(q=s\tau\). To derive the scale-conditioned MX-Gaussian CDF, we first use the rule of conditional probability, which yields 
 \begin{align}\label{eq:prob}
     \Pr(Z_i\leq q| \mxscale(\randomvec Y)=s)=\frac{\Pr(Z_i\leq q,\ \mxscale(\randomvec Y)=s)}{\Pr(\mxscale(\randomvec Y)=s)}. 
 \end{align}
 Next, the identity in \eqref{eq:mx-scale-cell} yields:
\begin{align}\label{eq:delta_scale}
    \Pr(Z_i\leq q,\ \mxscale(\randomvec Y)=s)=\Pr(Z_i\leq q,\ M < b(s) )-\Pr(Z_i\leq q,\ M<a(s)  ).
\end{align}


Since the magnitudes \(Z_j\) are independent, for \(c\geq0\) we have
\begin{equation}
\begin{aligned}
\Pr(Z_i\leq q,\ M<c)
&=\Pr\!\left(
    Z_i\leq\min\{q,c\},\ \max_{j\neq i}Z_j<c
  \right) =\Pr\!\left(Z_i\leq\min\{q,c\}\right)
  \prod_{j\neq i}\Pr(Z_j<c) \\
&=\stderf\!\left(\min\{q,c\}/\sqrt2\right)\stderf\!\left(c/\sqrt2\right)^{\dmx-1}.
\end{aligned}
\label{eq:mx-gaussian-proof-rectangle}
\end{equation}
In \eqref{eq:mx-gaussian-proof-rectangle}, the second step follows from the independence of the variables \(Z_j\), and the final step uses that \(Z_j=|Y_j|\) has a half-normal distribution, whose CDF for $z\geq 0$ equals $\stderf(z/\sqrt2)$.
Plugging \eqref{eq:mx-gaussian-proof-rectangle} into \eqref{eq:delta_scale} yields:
\begin{equation}\label{eq:mx-gaussian-proof-numerator}
\Pr(Z_i\leq q,\ \mxscale(\randomvec Y)=s)
=\stderf\!\left(\min\{q,b(s)\}/\sqrt2\right)
\stderf\!\left(b(s)/\sqrt2\right)^{\dmx-1}-\stderf\!\left(\min\{q,a(s)\}/\sqrt2\right)
\stderf\!\left(a(s)/\sqrt2\right)^{\dmx-1}.
\end{equation}
Finally, substituting \eqref{eq:mx-gaussian-proof-numerator} and \eqref{eq:scale_prob} into \eqref{eq:prob} gives, for all $\tau>0$,
\begin{equation}
\begin{aligned}
F(\tau;s)
&=\Pr\!\left(\lvert X_i\rvert\leq \tau\mid \mxscale(\randomvec Y)=s\right) \\
&=\Pr\!\left(Z_i\leq s \tau\mid \mxscale(\randomvec Y)=s\right) \\
&=\frac{\Pr(Z_i\leq q,\ \mxscale(\randomvec Y)=s)}{\Pr(\mxscale(\randomvec Y)=s)}
 =\frac{\Pr(Z_i\leq q,\ \mxscale(\randomvec Y)=s)}{\mxcellmass(s;\dmx)} \\
&=
\begin{cases}
\displaystyle
\frac{\stderf\!\left(q/\sqrt2\right)\mxcellmass(s;\dmx-1)}{\mxcellmass(s;\dmx)},
&0\leq q\leq a(s),\\[8pt]
\displaystyle
\frac{\stderf\!\left(q/\sqrt2\right)\stderf\!\left(b(s)/\sqrt2\right)^{\dmx-1}-\stderf\!\left(a(s)/\sqrt2\right)^{\dmx}}{\mxcellmass(s;\dmx)},
&a(s)<q<b(s),\\[8pt]
1,&q\geq b(s).
\end{cases}
\end{aligned}
\label{eq:mx-gaussian-proof-piecewise-cdf}
\end{equation}
In equation \eqref{eq:mx-gaussian-proof-piecewise-cdf}, the first step follows from the definition. The second step uses the relation \(\lvert X_i\rvert=Z_i/s\) on the event \(S=s\). Finally, we apply \eqref{eq:mx-gaussian-proof-numerator}, \eqref{eq:scale_prob}, and \eqref{eq:prob}.

\paragraph{Inverse CDF.}
The piecewise expression above proves
\cref{eq:mx-gaussian-conditional-cdf}.  Define the branch crossing in the
input domain as \(\tau^\star(s)=a(s)/s\), and define its CDF probability by
\begin{equation}
p^\star(s)
=F(\tau^\star(s);s)
=F(a(s)/s;s)
=\frac{\stderf\!\left(a(s)/\sqrt2\right)\mxcellmass(s;\dmx-1)}{\mxcellmass(s;\dmx)}.
\label{eq:mx-gaussian-proof-join}
\end{equation}
The inverse uses the lower CDF branch for
\(0<p\leq p^\star(s)\) and the upper branch for
\(p^\star(s)<p<1\).  Set \(\tau=F^{-1}(p;s)\), so that \(F(\tau;s)=p\). Solving the lower branch gives
\begin{equation}
\stderf\!\left(s \tau/\sqrt2\right)=\frac{\mxcellmass(s;\dmx)p}{\mxcellmass(s;\dmx-1)}.
\label{eq:mx-gaussian-proof-lower-inverse}
\end{equation}
Solving the upper branch gives
\begin{equation}
\stderf\!\left(s \tau/\sqrt2\right)=\frac{\mxcellmass(s;\dmx)p+\stderf\!\left(a(s)/\sqrt2\right)^{\dmx}}{\stderf\!\left(b(s)/\sqrt2\right)^{\dmx-1}}.
\label{eq:mx-gaussian-proof-upper-inverse}
\end{equation}
Applying $\stdierf$ and multiplying by $\sqrt2/s$ proves
\cref{eq:mx-gaussian-conditional-inverse}.  Both inverse branches place
\(\stderf\!\left(a(s)/\sqrt2\right)\) inside $\stdierf$ at \(p=p^\star(s)\), so the inverse
is continuous there.  
\end{proof}

%% file: section_files/clipped_mx_hash.tex
\section{Clipped MX-Hash}
\label{apx:clipped-mx-hash}
In this section, we derive the inverse CDF used by the clipped MX-Hash. We begin by giving a formal definition of the clipped MX-Gaussian distribution. Based on this definition, we then present its clipping probability. Finally, we derive its inverse CDF and define the hash.

\begin{definition}[Clipped MX-Gaussian]\label{def:clipped-mx-gaussian}
Let $\randomvec{Y}\sim\mathcal N(\mathbf 0,\matsym{I}_{\dmx})$, and fix a positive represented scale $s\in\mxscalegrid$. Let $\clipradius=\max\mxelementgrid>0$ be the maximum representable element magnitude. Define
\begin{equation}
\clippedcoord_i=\clipoperator(Y_i/s,-\clipradius,\clipradius),
\qquad i\in\{1,\ldots,\dmx\},
\label{eq:clipped-mx-gaussian-definition}
\end{equation}
where $\clipoperator(x,-\clipradius,\clipradius)=\max\{-\clipradius,\min\{x,\clipradius\}\}$. We call the distribution of the resulting vector the clipped MX-Gaussian distribution. Here $s$ is supplied externally; the Gaussian vector is not conditioned on the event $\mxscale(\randomvec Y)=s$.
\end{definition}

As a first step, we derive the clipping probability lemma.
\begin{lemma}[Clipping Probability]\label{lemma:clipped-mx-probability}
For each coordinate of the clipped MX-Gaussian vector, the probability at the maximum magnitude is
\begin{equation}
\Pr(|\clippedcoord_i|=\clipradius)
=1-\stderf(s\clipradius/\sqrt2).
\label{eq:clipped-mx-endpoint-mass}
\end{equation}
Each signed endpoint, $-\clipradius$ and $\clipradius$, has half this probability.
\end{lemma}
\begin{proof}
By the definition of clipping and the positivity of $s$, $|\clippedcoord_i|=\clipradius$ if and only if $|Y_i|\geq s\clipradius$. Since $Y_i\sim\mathcal N(0,1)$, the magnitude $|Y_i|$ has a half-normal distribution. Thus,
\[
\Pr(|\clippedcoord_i|=\clipradius)
=\Pr(|Y_i|\geq s\clipradius)
=1-\stderf(s\clipradius/\sqrt2).
\]
Symmetry of the Gaussian distribution gives equal probability to the two signed endpoints.
\end{proof}

Using this definition and Lemma~\ref{lemma:clipped-mx-probability}, we derive the inverse CDF in the following theorem:
\begin{theorem}[Clipped MX-Gaussian Inverse CDF]\label{thm:clipped-mx-inverse}
Let $\clippedcdf(\tau;s)=\Pr(|\clippedcoord_i|\leq\tau)$ be the magnitude CDF of the clipped MX-Gaussian distribution. For $0<p<1$, its generalized inverse is
\begin{equation}
\clippedcdf^{-1}(p;s)
=\min\left\{\frac{\sqrt2}{s}\stdierf(p),\clipradius\right\},
\label{eq:clipped-mx-inverse}
\end{equation}
where $\stderf$ is the standard error function and $\stdierf$ its inverse. The generalized inverse is defined by $\clippedcdf^{-1}(p;s)=\inf\{\tau\geq0:\clippedcdf(\tau;s)\geq p\}$.
\end{theorem}
\begin{proof}[Proof of \cref{thm:clipped-mx-inverse}]
For $0\leq\tau<\clipradius$, clipping preserves the event that the magnitude is at most $\tau$. Therefore,
\begin{equation}
\begin{aligned}
\clippedcdf(\tau;s)
&=\Pr(|\clippedcoord_i|\leq\tau)
=\Pr(|Y_i|/s\leq\tau)\\
&=\Pr(|Y_i|\leq s\tau)
=\stderf(s\tau/\sqrt2).
\end{aligned}
\label{eq:clipped-mx-interior-cdf}
\end{equation}
The second step follows from the definition of clipping, the third uses $s>0$, and the final step uses the half-normal CDF. The magnitude is nonnegative and never exceeds $\clipradius$, so
\begin{equation}
\clippedcdf(\tau;s)=
\begin{cases}
0,&\tau<0,\\
\stderf(s\tau/\sqrt2),&0\leq\tau<\clipradius,\\
1,&\tau\geq\clipradius.
\end{cases}
\label{eq:clipped-mx-cdf}
\end{equation}
The jump at $\clipradius$ equals the clipping probability in \eqref{eq:clipped-mx-endpoint-mass}.

\paragraph{Inverse CDF.}
Define the probability at the end of the continuous branch as
\begin{equation}
\clipcross(s)=\lim_{\tau\uparrow\clipradius}\clippedcdf(\tau;s)
=\stderf(s\clipradius/\sqrt2).
\label{eq:clipped-mx-crossing}
\end{equation}
For $0<p<\clipcross(s)$, set $\tau=\clippedcdf^{-1}(p;s)$. Solving the continuous branch gives
\begin{equation}
\stderf(s\tau/\sqrt2)=p,
\qquad
\tau=\frac{\sqrt2}{s}\stdierf(p).
\label{eq:clipped-mx-interior-inverse}
\end{equation}
For $\clipcross(s)\leq p<1$, every $\tau<\clipradius$ has $\clippedcdf(\tau;s)<p$, while $\clippedcdf(\clipradius;s)=1$. Thus the generalized inverse equals $\clipradius$. At $p=\clipcross(s)$, the continuous inverse also equals $\clipradius$. Combining the two branches proves \eqref{eq:clipped-mx-inverse}.
\end{proof}


%% file: section_files/efficient_lut.tex
\section{Efficient Look-up Table Implementation}
\label{app:efficient-lut}
In this section, we present an efficient approach for implementing the inverse CDF with lookup tables (LUTs). We describe LUT implementations of the MX-Gaussian and clipped MX-Gaussian inverse CDFs. Both implementations use table lookups, comparisons, multiplication, and addition, with interpolation factors precomputed where appropriate.

\subsection{MX-Gaussian}
For each supported scale $s$, we precompute the following coefficients for the MX-Gaussian inverse CDF: 
\begin{equation}
\begin{aligned}
\lutslopes_0(s)&=
 \frac{\mxcellmass(s;\dmx)}
 {\mxcellmass(s;\dmx-1)},
&
\lutslopes_1(s)&=
 \frac{\mxcellmass(s;\dmx)}{\stderf\!\left(b(s)/\sqrt2\right)^{\dmx-1}},
\\
\lutoffset(s)&=
 \frac{\stderf\!\left(a(s)/\sqrt2\right)^{\dmx}}{\stderf\!\left(b(s)/\sqrt2\right)^{\dmx-1}},
&
p^\star(s)&=
 \frac{\stderf\!\left(a(s)/\sqrt2\right)
 [\mxcellmass(s;\dmx-1)]}
 {\mxcellmass(s;\dmx)} .
\end{aligned}
\label{eq:lut-coefficients}
\end{equation}
We also store the prefactor $\sqrt2/s$. For an 8-bit scale encoding, we allocate 256 scale slots and store five FP32 values per slot. Next, we explain how to calculate the inverse CDF of the MX-Gaussian distribution.  Given the midpoint probability $p_\ell$ for output coordinate $\ell$ in \cref{eq:mx-hash-interior-probability}, we compute the inverse-error-function argument as follows: 
\begin{equation}
    \lutargument_\ell=\begin{cases}
        \lutslopes_0(s)p_\ell & p_\ell\le p^\star(s)\\
\lutslopes_1(s)p_\ell+\lutoffset(s)&  \text{otherwise }.
    \end{cases}
\end{equation}
We approximate the inverse error function using tabulated values and linear interpolation. In particular, let $\lutknots_0,\ldots,\lutknots_{255}$ denote the set of points in the LUT. Because the input to the inverse error function spans 0 to 1, we set $\lutknots_0=0$ and $\lutknots_{255}=1$, and linearly distribute the intermediate points between these endpoints. We construct the inverse-error-function lookup table as $\lutvalues_j=\stdierf(\lutknots_j)$. We first approximate $\stdierf(\lutargument_\ell)$ by $\Lambda(\lutargument_\ell)$. The approximate magnitude quantile is $(\sqrt2/s)\Lambda(\lutargument_\ell)$; applying the sign and element quantizer gives the approximate hash output. For $\lutknots_j\le\lutargument_\ell\le\lutknots_{j+1}$, these quantities are evaluated as follows:
\begin{equation}
\Lambda(\lutargument_\ell)=
\lutvalues_j+
\frac{\lutargument_\ell-\lutknots_j}{\lutknots_{j+1}-\lutknots_j}
(\lutvalues_{j+1}-\lutvalues_j).
\label{eq:lut-interpolation}
\end{equation}
\begin{equation}
\widehat h_{\hashparams,\ell}(u;s)=
Q_E\!\left(
(-1)^{\mxsignbit_\ell}\frac{\sqrt2\,\Lambda(\lutargument_\ell)}{s}
\right).
\label{eq:lut-hash-output}
\end{equation}

The two 256-entry FP32 arrays require 2\,KiB in total. Five FP32 coefficients per scale slot require another 5\,KiB for 256 slots, giving 7\,KiB.


\subsection{Clipped MX-Gaussian}
\label{app:clipped-efficient-lut}
For each supported scale $s$, we precompute the prefactor $\sqrt2/s$. For an 8-bit scale encoding, we allocate 256 scale slots and store one FP32 value per slot. We reuse the inverse-error-function LUT and linear interpolation from the MX-Gaussian implementation in \cref{eq:lut-interpolation}. By \cref{eq:clipped-mx-inverse}, the inverse-error-function argument is $p_\ell$ itself. Applying clipping, the sign, and the element quantizer gives
\begin{equation}
\widehat{\clippedhash}_{\hashparams,\ell}(u;s)
=Q_E\!\left((-1)^{\mxsignbit_\ell}
\min\!\left\{\frac{\sqrt2}{s}\Lambda(p_\ell),\clipradius\right\}\right).
\label{eq:clipped-lut-hash}
\end{equation}
The shared two 256-entry FP32 arrays require 2\,KiB in total. One FP32 prefactor per scale slot requires another 1\,KiB for 256 slots, giving 3\,KiB for a standalone implementation or 1\,KiB additional storage when the LUT is shared.

\subsection{Efficient Hash Implementation Results}

\label{apx:efficient-hash}

\begin{figure}[H]
\centering
\includegraphics[width=3.3in]{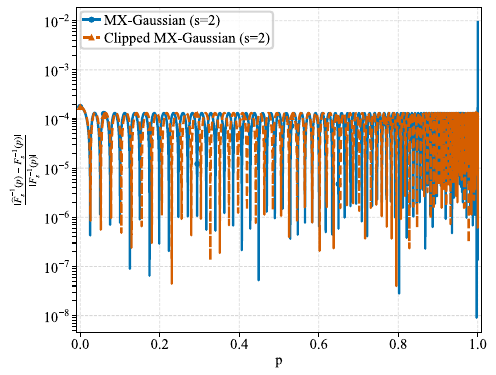}
\caption{Sampled relative inverse-CDF error at $s=2$ with linear interpolation; zero errors are omitted.}
\label{fig:lut-relative-error}
\end{figure}

Figure~\ref{fig:lut-relative-error}\footnote{This diagnostic uses 256 finite, nonuniform inverse-erf knots and FP64 arithmetic, rather than the uniform FP32 scheme above; arguments outside the table use the full inverse CDF (35 MX-Gaussian samples; none for the clipped distribution).} shows sampled relative inverse-CDF errors before element rounding at $s=2$ (FP4 E2M1, block size 32). The relative error remains below $1\%$ across the sampled probabilities for both distributions, indicating close agreement with the full inverse-CDF reference at this scale. Next, we compare perplexity using the reference and efficient implementations.

Table~\ref{tab:efficient-hash} compares the reference and efficient
hash implementations of \MN{} at matched effective storage budgets.
UE8M0 uses initial scale selection, while UE5M3 uses joint optimization.
Reference results are the reported main results in Section~\ref{sec:experiments}.
The results show that the efficient hash implementation maintains perplexity close to the original across both scale formats, improving results on Qwen3-8B while performing slightly worse on Llama-3.1-8B-Instruct. These results suggest that the computational simplification largely preserves quantization quality.

\begin{table}[H]
\centering
\caption{Perplexity of \MN{} with and without efficient hashing.
Bold marks the best result between the two implementations for each
model and scale format. BF16 is excluded from this comparison.
UE8M0 uses initial scale selection and UE5M3 uses joint optimization.}
\label{tab:efficient-hash}
\begingroup
\footnotesize
\setlength{\tabcolsep}{3pt}
\setlength{\arrayrulewidth}{0.3pt}
\renewcommand{\arraystretch}{1.35}
\resizebox{0.65\textwidth}{!}{%
\begin{tabular}{@{}clc|c|c@{}}
\hline
\multirow{2}{*}{\shortstack{Block-scale\\format}}
& \multirow{2}{*}{Variant}
& \multirow{2}{*}{\shortstack{Effective\\Bit}}
& \textbf{Llama-3.1-8B-Instruct}
& \textbf{Qwen3-8B} \\
& & &
Perplexity $\downarrow$ & Perplexity $\downarrow$ \\
\hline
---
& BF16
& 16
& 7.170
& 9.750 \\
\hline
\multirow{2}{*}{UE8M0}
& \MN{}
& 1.25
& \textbf{11.957}
& 15.072 \\
& \MN{} efficient hash
& 1.25
& 11.980
& \textbf{14.831} \\
\hline
\multirow{2}{*}{UE5M3}
& \MN{}
& 1.25
& \textbf{11.630}
& 14.657 \\
& \MN{} efficient hash
& 1.25
& 11.712
& \textbf{14.537} \\
\hline
\end{tabular}%
}
\endgroup
\end{table}

%% file: section_files/joint_scale_search.tex
\section{Relaxed Joint Scale Search}
\label{app:joint-scale-search}

In this section, we derive the scale search algorithm introduced in \cref{sec:joint-scale-search}.
Let $\y_{b,j},\widehat{\x}_{b,j}\in\R^V$ denote the $j$th groups of $V$ coordinates in $\y_b$ and $\widehat{\x}_b$, respectively.
With $s_b$ fixed, the prefix reconstruction error for a candidate path over the first $m\in\{1,\ldots,\dmx/V\}$ emissions of block $b$ is
\begin{equation}
E_{b,m}(s_b) = \sum_{j=1}^{m}\left\|\y_{b,j} - s_b\widehat{\x}_{b,j}\right\|_2^2.
\end{equation}
At $m=\dmx/V$, this recovers the full block reconstruction error of \cref{eq:mx-reconstruction-error}.
The \emph{fixed-scale} beam selects the $B$ lowest-cost candidates according to the accumulated error of completed blocks plus $E_{b,m}(s_b)$.

To construct a joint scale search, we first relax the represented block scale $s_b$ to a continuous variable $a_b>0$, with $s_b=Q_\alpha(a_b)$.
Let $\heuristicscale_b=Q_\alpha(\mxrefscale_b)$ denote the fixed reference scale selected before search.
Writing $E_{b,m}(a_b)$ for the same prefix error evaluated at the continuous scale, define the regularized squared error
\begin{align}
E_{b,m}^{\mathrm{reg}}(a_b)
&=E_{b,m}(a_b)
+\lambda\!\left(a_b-\heuristicscale_b\right)^2,
\label{eq:mx-relaxed-prefix-objective}
\end{align}
where $\lambda\geq0$ controls the regularization strength.
This encourages scales that are suited to the whole block, when $m<\dmx/V$.
For example, if $m=V=1$ and $\lambda=0$, $a_b$ can be chosen to reconstruct $\y_{b,1}$ for any nonzero emission $\widehat{\x}_{b,1}$ of the same sign.
Such paths can receive low prefix costs despite scale choices that lead to large errors later in the block, potentially evicting better paths from the finite beam.

To support this joint search, we use Clipped MX-Hash (Appendix~\ref{apx:clipped-mx-hash}), whose quantized outputs $\widehat{\x}_{b,j}$ remain fixed as the block scale varies.
The regularized objective is therefore quadratic in $a_b$.
Setting $\partial E_{b,m}^{\mathrm{reg}}(a_b)/\partial a_b=0$ gives
\begin{equation}
a_{b,m}^{*}
=\frac{\phi_m(\y_b,\widehat{\x}_b)+\lambda\heuristicscale_b}
{\phi_m(\widehat{\x}_b,\widehat{\x}_b)+\lambda},
\label{eq:mx-relaxed-scale-update}
\end{equation}
provided the denominator and fitted scale are positive.
Here, the prefix inner product is
\begin{equation}
\phi_m(\mathbf v,\mathbf w)
=\sum_{j=1}^{m}\langle\mathbf v_j,\mathbf w_j\rangle.
\end{equation}
Both prefix inner products accumulate with each emission, so scale refitting takes constant work per candidate.
We evaluate $E_{b,m}^{\mathrm{reg}}(s_{b,m})$ at the rounded scale $s_{b,m}=Q_\alpha(a_{b,m}^{*})$ and add the accumulated error of completed blocks to rank candidates.
For the final step of each MX block, we set $\lambda=0$, since the full block reconstruction error is now available. We then refit and quantize the scale before calculating the final block error.

This \emph{relaxed joint scale} beam search selects the trellis path and block scales in a single beam search.
Algorithm~\ref{alg:trellis_beam_encoding} details the joint-scale encoding procedure.

%% file: section_files/additional_results.tex
\raggedbottom

\section{Additional Results}

\subsection{Below 1-bit PPL Results}
\label{apx:below_1bit_res}

Table~\ref{tab:below-one-bit-ppl} extends the WikiText-2 evaluation to storage budgets below one effective bit per element, with VecInfer at one bit as a reference.
At below 1 effective bits per element, the MX trellis variants achieve lower perplexity than the compared baselines on both models under both scale formats.
The UE5M3 configuration with joint scale search further reduces perplexity and, even at approximately $0.93$ effective bits per element, outperforms NSNQuant at a slight higher effective bit on both models.

\begin{table}[H]
\centering
\color{black}
\caption{Below-1-bit MX KV-cache quantization perplexity. 
Effective Bit denotes mean effective bits per element using dynamic storage accounting. Within each block-scale format and model, dark blue cells mark the lowest perplexity and light blue cells mark the second-lowest perplexity among the displayed quantized methods.}
\label{tab:below-one-bit-ppl}
\begingroup
\footnotesize
\setlength{\tabcolsep}{3pt}
\setlength{\arrayrulewidth}{0.3pt}
\renewcommand{\arraystretch}{1.35}
\resizebox{0.6\textwidth}{!}{%
\begin{tabular}{@{}ccc|c|c@{}}
\hline
\shortstack{Block-scale\\format} & Variant & \shortstack{Effective\\Bit} & \textbf{Llama-3.1-8B-Instruct} & \textbf{Qwen3-8B} \\
\hline
--- & BF16 & 16.00 & 7.170 & 9.750 \\
\hline
\multirow{4}{*}{UE8M0} & VecInfer & 1.00 & 693.850 & 409.782 \\
 & NSNQuant & 0.99 & \cellcolor[HTML]{DDEBF7}\color{black} 52.257 & 80.740 \\
 & \TableMN{} & 0.99 & \cellcolor[HTML]{9DC3E6}\color{black} 38.315 & \cellcolor[HTML]{9DC3E6}\color{black} 57.616 \\
 & \TableMN{} & 0.93 & 57.510 & \cellcolor[HTML]{DDEBF7}\color{black} 79.069 \\
\hline
\multirow{4}{*}{UE5M3} & VecInfer & 1.00 & 694.108 & 324.561 \\
 & NSNQuant & 0.99 & 54.845 & 87.735 \\
 & \TableMN{} & 0.99 & \cellcolor[HTML]{9DC3E6}\color{black} 32.136 & \cellcolor[HTML]{9DC3E6}\color{black} 42.248 \\
 & \TableMN{} & 0.93 & \cellcolor[HTML]{DDEBF7}\color{black} 45.496 & \cellcolor[HTML]{DDEBF7}\color{black} 59.141 \\
\hline
\end{tabular}
}
\endgroup
\end{table}

\subsection{\MN{} Integration with VecInfer}
\label{apx:vec-comparison}

In Table~\ref{tab:vecinfer-qtip-lowbit} we evaluate \MN{} as a drop-in replacement for VecInfer's vector quantizer, retaining its preprocessing.
At matched storage budgets, \MN{} achieves lower perplexity and higher LongBench and GSM8K scores on both models across both scale formats.

\begin{table}[H]
\centering
\color{black}
\caption{VecInfer with and without \MN{} at $1.25$ effective bits per element. Blue marks the best result per model, scale format, and metric; both UE8M0 Llama GSM8K results tie at the displayed precision.}
\label{tab:vecinfer-qtip-lowbit}
\begingroup
\footnotesize
\setlength{\tabcolsep}{3pt}
\setlength{\arrayrulewidth}{0.3pt}
\renewcommand{\arraystretch}{1.35}
\resizebox{\textwidth}{!}{%
\begin{tabular}{@{}clc|ccc|ccc@{}}
\hline
\multirow{2}{*}{\shortstack{Block-scale\\format}}
& \multirow{2}{*}{Variant}
& \multirow{2}{*}{\shortstack{Effective\\Bit}}
& \multicolumn{3}{c|}{\textbf{Llama-3.1-8B-Instruct}}
& \multicolumn{3}{c}{\textbf{Qwen3-8B}} \\
& & &
Perplexity $\downarrow$ & LongBench $\uparrow$ & GSM8K $\uparrow$ &
Perplexity $\downarrow$ & LongBench $\uparrow$ & GSM8K $\uparrow$ \\
\hline
\multirow{2}{*}{UE8M0}
& VecInfer
& 1.25
& 258.713
& 1.20
& \cellcolor[HTML]{9DC3E6}\color{black} 1.97
& 35.042
& 6.48
& 2.20 \\
& \textbf{VecInfer + \MN{} (Ours)}
& 1.25
& \cellcolor[HTML]{9DC3E6}\color{black} 188.996
& \cellcolor[HTML]{9DC3E6}\color{black} 2.11
& \cellcolor[HTML]{9DC3E6}\color{black} 1.97
& \cellcolor[HTML]{9DC3E6}\color{black} 18.967
& \cellcolor[HTML]{9DC3E6}\color{black} 11.46
& \cellcolor[HTML]{9DC3E6}\color{black} 3.41 \\
\hline
\multirow{2}{*}{UE5M3}
& VecInfer
& 1.25
& 232.335
& 1.22
& 1.90
& 30.791
& 11.43
& 2.50 \\
& \textbf{VecInfer + \MN{} (Ours)}
& 1.25
& \cellcolor[HTML]{9DC3E6}\color{black} 69.729
& \cellcolor[HTML]{9DC3E6}\color{black} 3.96
& \cellcolor[HTML]{9DC3E6}\color{black} 1.97
& \cellcolor[HTML]{9DC3E6}\color{black} 15.494
& \cellcolor[HTML]{9DC3E6}\color{black} 19.32
& \cellcolor[HTML]{9DC3E6}\color{black} 9.40 \\
\hline
\end{tabular}%
}
\endgroup
\end{table}

\subsection{Full LongBench Per Task Results}
\label{apx:longbench_full}

Tables~\ref{tab:longbench-l318bi} and~\ref{tab:longbench-q38} provide the individual scores underlying the LongBench averages in the main results, covering question answering, summarization, and classification.
The results show that the benefits of \MN{} extend across multiple task types, with broad gains for the lower-bit UE5M3 and NVFP4 configurations on Llama-3.1-8B-Instruct.

\begin{table}[H]
\centering
\color{black}
\caption{LongBench task scores for Llama-3.1-8B-Instruct ($\uparrow$).
Bold marks the best per scale and bit section, excluding historical BF16.}
\label{tab:longbench-l318bi}
\begingroup
\footnotesize
\setlength{\tabcolsep}{3pt}
\setlength{\arrayrulewidth}{0.3pt}
\renewcommand{\arraystretch}{1.35}
\resizebox{0.8\textwidth}{!}{%
\begin{tabular}{@{}clc|ccccccc@{}}
\hline
\shortstack{Block-scale\\format} & Variant & \shortstack{Effective\\Bit} & GovReport & Qasper & QMSum & SAMSum & MultiNews & TriviaQA & TREC \\
\hline
--- & BF16 & 16.00 & 34.52 & 44.68 & 25.20 & 43.92 & 27.02 & 91.65 & 72.50 \\
\hline
\multirow{3}{*}{UE8M0}
 & VecInfer & 2.25
 & 29.27 & 34.39 & 22.03
 & 39.01 & 26.27 & 81.72 & 63.00 \\
 & NSNQuant & 2.24
 & 33.70 & 42.96 & 24.86
 & \textbf{43.81} & 26.40 & \textbf{91.12} & 67.50 \\
 & \TableMN{} & 2.25
 & \textbf{34.54} & \textbf{43.91} & \textbf{24.95}
 & 43.37 & \textbf{26.67} & 89.23 & \textbf{70.50} \\
\hline
\multirow{3}{*}{UE5M3} & VecInfer & 2.25
 & 30.41 & 39.93 & 22.42 & 38.64 & 26.41 & 79.27 & 68.00 \\
 & NSNQuant & 2.24 & 33.30 & \textbf{45.38} & 25.33 & 43.83 & 26.57 & 88.93 & 70.00 \\
 & \TableMN{} & 2.25 & \textbf{34.15} & 45.23 & \textbf{26.11} & \textbf{43.99} & \textbf{26.58} & \textbf{90.38} & \textbf{72.00} \\
\hline
\multirow{3}{*}{NVFP4} & VecInfer & 2.25 & 29.47 & 35.51 & 22.29 & 38.11 & 26.43 & 83.56 & 63.50 \\
 & NSNQuant & 2.24 & 33.83 & \textbf{45.86} & 25.09 & \textbf{43.64} & \textbf{27.13} & 89.70 & 70.00 \\
 & \TableMN{} & 2.25 & \textbf{34.64} & 43.62 & \textbf{25.57} & 42.79 & 26.10 & \textbf{90.90} & \textbf{71.00} \\
\hline
\hline
\multirow{3}{*}{UE8M0} & VecInfer & 1.25
 & 1.11 & 0.73 & 1.94 & 1.74 & 1.52 & 0.34 & 1.00 \\
 & NSNQuant & 1.24 & 27.08 & \textbf{25.06} & \textbf{21.98} & \textbf{37.38} & 24.77 & \textbf{74.83} & \textbf{56.50} \\
 & \TableMN{} & 1.25
 & \textbf{27.49} & \textbf{28.60} & \textbf{22.21}
 & 34.70 & \textbf{25.49} & 67.57 & 44.50 \\
\hline
\multirow{3}{*}{UE5M3} & VecInfer & 1.25
 & 1.15 & 0.66 & 2.05 & 2.18 & 1.26 & 0.23 & 1.00 \\
 & NSNQuant & 1.24 & 28.07 & 26.38 & \textbf{22.67} & 36.19 & 25.10 & 72.13 & \textbf{56.50} \\
  & \TableMN{} & 1.25
 & \textbf{29.95} & \textbf{29.44} & 22.46
 & \textbf{40.23} & \textbf{25.59} & \textbf{79.40} & 55.50 \\
\hline
\multirow{3}{*}{NVFP4} & VecInfer & 1.25 & 1.22 & 0.80 & 2.17 & 2.35 & 1.74 & 0.72 & 2.50 \\
 & NSNQuant & 1.24 & 27.53 & 21.95 & 22.30 & 38.48 & 24.96 & 71.19 & \textbf{59.00} \\
 & \TableMN{} & 1.25 & \textbf{30.06} & \textbf{32.66} & \textbf{22.79} & \textbf{39.68} & \textbf{26.02} & \textbf{82.43} & 53.50 \\
\hline
\end{tabular}%
}
\endgroup
\end{table}

\begin{table}[H]
\centering
\color{black}
\caption{LongBench task scores for Qwen3-8B ($\uparrow$).
Bold marks the best per scale and bit section, excluding historical BF16.
UE8M0/UE5M3 VecInfer uses MX-unaware quantization; NVFP4 retains the original setup.
All \MN{} rows use beam 128.}
\label{tab:longbench-q38}
\begingroup
\footnotesize
\setlength{\tabcolsep}{3pt}
\setlength{\arrayrulewidth}{0.3pt}
\renewcommand{\arraystretch}{1.35}
\resizebox{0.8\textwidth}{!}{%
\begin{tabular}{@{}clc|ccccccc@{}}
\hline
\shortstack{Block-scale\\format} & Variant & \shortstack{Effective\\Bit} & GovReport & Qasper & QMSum & SAMSum & MultiNews & TriviaQA & TREC \\
\hline
--- & BF16 & 16.00 & 27.86 & 12.03 & 18.06 & 44.54 & 22.77 & 90.71 & 71.50 \\
\hline
\multirow{3}{*}{UE8M0}
 & VecInfer & 2.25
 & 26.08 & \textbf{12.88} & \textbf{18.92}
 & 39.30 & 21.86 & 78.53 & 69.50 \\
 & NSNQuant & 2.24
 & 26.62 & 11.62 & 17.64
 & \textbf{43.24} & 22.58 & 88.55 & \textbf{72.50} \\
 & \TableMN{} & 2.25
 & \textbf{27.41} & 11.42 & 17.85
 & 42.35 & \textbf{22.62} & \textbf{89.56} & 72.00 \\
\hline
\multirow{3}{*}{UE5M3} & VecInfer & 2.25
 & 26.34 & \textbf{12.54} & \textbf{18.44}
 & 39.09 & 21.84 & 81.07 & 71.00 \\
 & NSNQuant & 2.24 & 27.10 & 11.33 & 17.80 & 44.36 & \textbf{22.75} & 89.81 & \textbf{75.00} \\
 & \TableMN{} & 2.25 & \textbf{27.76} & 11.34 & 17.81 & \textbf{44.45} & 22.42 & \textbf{90.06} & 70.50 \\
\hline
\multirow{3}{*}{NVFP4} & VecInfer & 2.25 & 26.29 & 13.42 & \textbf{18.72} & 39.92 & 22.11 & 82.23 & 74.50 \\
 & NSNQuant & 2.24 & 27.05 & 11.50 & 17.32 & 43.39 & 22.27 & \textbf{89.47} & 69.00 \\
 & \TableMN{} & 2.25 & \textbf{27.69} & \textbf{11.55} & 17.74 & \textbf{44.41} & \textbf{22.39} & 88.15 & \textbf{72.50} \\
\hline
\hline
\multirow{3}{*}{UE8M0} & VecInfer & 1.25
 & 7.96 & 2.19 & 11.22 & 4.63 & 10.86 & 2.98 & 5.50 \\
 & NSNQuant & 1.24 & \textbf{23.10} & 8.47 & \textbf{16.75} & \textbf{30.63} & 18.53 & \textbf{69.72} & 49.50 \\
  & \TableMN{} & 1.25
 & 22.89 & \textbf{8.81} & 16.38
 & 29.22 & \textbf{20.66} & 64.51 & \textbf{59.50} \\
\hline
\multirow{3}{*}{UE5M3} & VecInfer & 1.25
 & 11.26 & 4.04 & 12.49 & 4.12 & 14.77 & 2.83 & 30.50 \\
 & NSNQuant & 1.24 & 22.79 & 9.00 & \textbf{17.27} & 30.62 & 18.39 & 71.56 & 55.50 \\
  & \TableMN{} & 1.25
 & \textbf{23.23} & \textbf{9.68} & 16.25
 & \textbf{31.12} & \textbf{20.06} & \textbf{71.62} & \textbf{62.00} \\
\hline
\multirow{3}{*}{NVFP4} & VecInfer & 1.25 & 10.41 & 3.26 & 12.24 & 3.89 & 14.90 & 4.78 & 22.00 \\
 & NSNQuant & 1.24 & 23.39 & 8.35 & \textbf{16.85} & 29.12 & 18.21 & \textbf{69.43} & 54.00 \\
 & \TableMN{} & 1.25 & \textbf{23.40} & \textbf{8.87} & 16.73 & \textbf{34.36} & \textbf{20.67} & 66.40 & \textbf{61.00} \\
\hline
\end{tabular}%
}
\endgroup
\end{table}